\documentclass[lettersize,journal]{IEEEtran}
\usepackage{amsmath,amsfonts}
\usepackage{algorithmic}
\usepackage{algorithm}

\usepackage{array}
\usepackage{textcomp}
\usepackage{stfloats}
\usepackage{url}
\usepackage{verbatim}
\usepackage{graphicx}
\newcommand{\methodname}{LeapAD}
\newcommand{\newmethodname}{LeapVAD}

\newcommand{\fastsystem}{Heuristic Process}
\newcommand{\slowsystem}{Analytic Process}

\usepackage[utf8]{inputenc} 
\usepackage[T1]{fontenc}    
\usepackage{hyperref}       
\usepackage{url}            
\usepackage{booktabs}       
\usepackage{diagbox}
\usepackage{amsfonts}       
\usepackage{nicefrac}       
\usepackage{enumitem}
\usepackage{microtype}      
\usepackage[table,xcdraw]{xcolor}
\usepackage{colortbl}
\usepackage{multirow}
\usepackage{comment}
\usepackage{amsmath}
\usepackage{wrapfig}
\usepackage{makecell}
\usepackage{arydshln}
\usepackage{multicol}
\usepackage{tabularx}

\usepackage{svg}

\usepackage{subcaption}  

\definecolor{MyDarkBlue}{rgb}{0,0.08,1}
\definecolor{MyDarkGreen}{rgb}{0.02,0.6,0.02}
\definecolor{MyDarkRed}{rgb}{0.8,0.02,0.02}
\definecolor{MyDarkOrange}{rgb}{0.40,0.2,0.02}
\definecolor{MyPurple}{RGB}{111,0,255}
\definecolor{MyRed}{rgb}{1.0,0.0,0.0}
\definecolor{MyGold}{rgb}{0.75,0.6,0.12}
\definecolor{MyDarkgray}{rgb}{0.66, 0.66, 0.66}

\newcommand{\gray}[1]{\textcolor[rgb]{0.5,0.5,0.5}{#1}}

\def\BibTeX{{\rm B\kern-.05em{\sc i\kern-.025em b}\kern-.08em
    T\kern-.1667em\lower.7ex\hbox{E}\kern-.125emX}}
\usepackage{balance}
\begin{document}
\title{LeapVAD: A Leap in Autonomous Driving via Cognitive Perception and Dual-Process Thinking}
\author{Yukai Ma$^{1,2}$, Tiantian Wei$^{3,2}$,  Naiting Zhong$^{4,2}$, Jianbiao Mei$^{1,2}$, Tao Hu$^{5,2}$,\\Licheng Wen$^{2}$, Xuemeng Yang$^{2}$, Botian Shi$^{2,*}$, Yong Liu$^{1,*}$ 
\thanks{$^{1}$The authors are with the Institute of Cyber-Systems and Control, Zhejiang University, Hangzhou, China.}%
\thanks{$^{2}$The authors are with the Shanghai Artificial Intelligence Laboratory, Shanghai, China.}%
\thanks{$^{3}$The author is with TUM School of Engineering and Design, Technical University of Munich, Munich, Germany}%
\thanks{$^{4}$The author is with Tongji University}%
\thanks{$^{5}$The author is with Science Island Branch of Graduate School, University of Science and Technology of China}%

\thanks{$^{*}$Botian Shi and Yong Liu are the corresponding authors (Email: shibotian@pjlab.org.cn; yongliu@iipc.zju.edu.cn)}%
}

\maketitle

\begin{abstract}
While autonomous driving technology has made remarkable strides, data-driven approaches still struggle with complex scenarios due to their limited reasoning capabilities. Meanwhile, knowledge-driven autonomous driving systems have evolved considerably with the popularization of visual language models. In this paper, we propose {\newmethodname}, a novel method based on cognitive perception and dual-process thinking. Our approach implements a human-attentional mechanism to identify and focus on critical traffic elements that influence driving decisions. By characterizing these objects through comprehensive attributes - including appearance, motion patterns, and associated risks - LeapVAD achieves more effective environmental representation and streamlines the decision-making process.
Furthermore, {\newmethodname} incorporates an innovative dual-process decision-making module miming the human-driving learning process. The system consists of an {\slowsystem} (System-II) that accumulates driving experience through logical reasoning and a {\fastsystem} (System-I) that refines this knowledge via fine-tuning and few-shot learning. {\newmethodname} also includes reflective mechanisms and a growing memory bank, enabling it to learn from past mistakes and continuously improve its performance in a closed-loop environment. 
To enhance efficiency, we develop a scene encoder network that generates compact scene representations for rapid retrieval of relevant driving experiences.
Extensive evaluations conducted on two leading autonomous driving simulators, CARLA and DriveArena, demonstrate that {\newmethodname} achieves superior performance compared to camera-only approaches despite limited training data. Comprehensive ablation studies further emphasize its effectiveness in continuous learning and domain adaptation.
Project page: \url{https://pjlab-adg.github.io/LeapVAD/}.
\end{abstract}


\section{Introduction}
Since the early 21st century, humans have been exploring the use of computer algorithms to replace human drivers. 

Recent data-driven approaches~\cite{yin2021center,li2022bevformer,liu2023bevfusion} have achieved remarkable success. However, they often rely heavily on the distribution of training data, which can result in a superficial understanding of underlying semantics and lead to misinterpretations in complex or unfamiliar scenarios. Data-driven approaches typically generalize observed patterns without inferential capabilities, limiting their performance to the scope of annotated data. Consequently, there is a pressing need for a system that can reason beyond the boundaries of its training data and imitate human cognitive processes. Several knowledge-based methods~\cite{shao2023lmdrive, mao2023language, yuan2024rag, tian2024drivevlm} employ large language models (LLMs) and vision language models (VLMs) as driving agents, marking a significant step towards more advanced autonomous systems. However, current evaluation methods for these approaches, such as open-loop testing, fall short of capturing the dynamic interactions between the self-driving car and its environment~\cite{li2023ego}. As a result, the responsiveness and adaptability of driving agents may not be adequately assessed, highlighting the need for more comprehensive evaluation methodologies.

Human learning to drive involves continuous interaction in closed-loop environments, where decisions are made based on surroundings and feedback is received. According to dual-process theory~\cite{kahneman2011fast, evans2013dual, wason1974dual}, human intelligence operates on two systems: 1) \emph{System-I} (i.e., \fastsystem, which is characterized by being fast, automatic, and empirical) and 2) \emph{System-II} (i.e., \slowsystem, which is characterized by being slow, rational, and logical). This dual-process thinking is evident as novice drivers transition to experienced ones. Initially, they depend on common sense, but with training, they develop skills through trial and error and rational analysis (\slowsystem), leading to muscle memory that allows for quick, instinctive reactions in familiar situations (\fastsystem). Even after obtaining a driver’s license, individuals continue to learn from experiences and accidents to improve their driving skills.

Our initial work in \cite{mei2024continuously} introduced a dual-process driving system inspired by the human attention mechanism. This system aimed to achieve knowledge-driven autonomous driving with limited training data, emulating the human ability to learn from experience and refine skills over time. While the previous method demonstrated promising performance, it had several limitations, such as supporting only single-frame image inputs and lacking precise motion prediction for traffic participants. To overcome these challenges, we propose an enhanced version named {\newmethodname}. Specifically, the algorithm has been extended to support multi-view and multi-frame inputs. Furthermore, we enhance the scene encoder to extract scene tokens that are more closely related to driving actions, thus improving the accuracy of the overall system.

Building on the dual-process continuous learning framework, this paper presents the following new contributions:

\begin{itemize}[align=right,itemindent=0em,labelsep=2pt,labelwidth=1em,leftmargin=*,itemsep=0em]

\item \textbf{Temporal Scene Understanding.} We extend our approach to handle multi-frame inputs, which provide a more comprehensive understanding of key traffic objects by offering attributes such as velocity and motion trends across multiple frames.

\item \textbf{Efficient Retrieval Mechanism.} We propose a Scene Encoder that effectively captures scene tokens pertinent to anticipated driving actions, facilitating the acquisition of few-shot samples. Compared to text embedding, our method achieves higher precision in encoding scene tokens.

\item \textbf{Knowledge Transferability in Domain Adaptation.}
{\newmethodname} utilizes {\slowsystem} and reflection mechanisms to compile a memory bank of transferable experiences. Our results show that the knowledge stored in this memory bank can be applied not only to various towns within the same domain but also to different domains, such as high-fidelity environments.

\item \textbf{Extended Validation.} We validate our approach across a broader range of experimental setups and scenarios. Ultimately, we improved our driving score on the CARLA~\cite{dosovitskiy2017carla} Town05 short and long benchmarks by 5.3\% and 42.6\%, respectively, compared to our prior results~\cite{mei2024continuously}. Moreover, we achieved the best performance on DriveArena~\cite{yang2024drivearena} by utilizing the memory bank selected in CARLA.
\end{itemize}

\begin{algorithm}[t]
\caption{\newmethodname}\label{alg:alg1}
\begin{algorithmic}
\STATE \textbf{Input:}
    \STATE \hspace{0.2cm} System Prompt $P_s$
    \STATE \hspace{0.2cm} Reflection Prompt $P_r$
    \STATE \hspace{0.2cm} Traffic Rules $P_t$
    \STATE \hspace{0.2cm} Images $\mathbf{I}$ from ego car
    \STATE \hspace{0.2cm} Current ego state $\mathbf{A}$
    \STATE \hspace{0.2cm} Navigation information $N$ from LimSim~\cite{wen2023limsim}
\STATE \textbf{Output:} 
\STATE \hspace{0.2cm} Decisions $S$ for LimSim~\cite{wen2023limsim}
\STATE
\STATE {\textsc{Analytic Process~(Section~\ref{method:slow})}}
\FOR{$i < \text{length(sequence)}$}
    \STATE $D_i \gets \text{VLM}(P_s, \mathbf{I}_i)$ \hfill {\gray{\% Section~\ref{method:vlm}}}
    \STATE $R_i, S_i \gets \text{GPT}(P_t, D_i, N_i, \mathbf{A}_i)$ \hfill {\gray{\% reasoned decisions}}
    \STATE $\mathbf{t}_i \gets \mathbf{E}(\mathbf{I}_i, \mathbf{A}_i)$ \hfill {\gray{\% Section~\ref{sec:scenetoken}}}
    \STATE $\mathbf{B} \gets \mathbf{B} \cup \{\mathbf{t}_i, D_i, R_i, S_i\}$ \hfill {\gray{\% update memory bank}}
    \STATE \textbf{send} $S_i$ \hfill {\gray{\% send meta actions to low-level controller}}
\ENDFOR
\STATE $\mathbf{B} \gets \text{revise}(\mathbf{B})$ \hfill {\gray{\% check and update}}

\STATE
\STATE {\textsc{Heuristic Process~(Section~\ref{method:fast})}}
\FOR{$j < \text{length(sequence)}$}
    \STATE $D_j \gets \text{VLM}(P_s, \mathbf{I}_j)$
    \STATE $\mathbf{t}_j \gets \mathbf{E}(\mathbf{I}_j, \mathbf{A}_j)$
    \STATE $\{\boldsymbol{t}_i, D_i, R_i, S_i\}^{k-1}_{i=0} \gets \text{TopK}(\boldsymbol{B},k) $ \hfill {\gray{{\% top-k samples}}}
    \STATE $R_j,S_j \gets \text{LLM}(P_t, \{D_i, R_i, S_i\}^{k-1}_{i=0}, D_j, N_j, \mathbf{A}_j)$
    \STATE $\mathbf{Q} \gets \mathbf{Q} \cup \{\mathbf{t}_j, D_j, R_j, S_j\}$ \hfill {\gray{\% historical queues}}
    \IF{$\text{length}(\mathbf{Q}) > m$}
        \STATE $\text{dequeue}(\mathbf{Q})$ \hfill {\gray{\% remove oldest element}}
    \ENDIF
    \STATE \textbf{send} $S_j$
\ENDFOR
\STATE
\STATE {\textsc{Reflection Mechanism~(Section~\ref{sec:reflection})}}
\IF{$\text{detect\_accident}() = \text{True}$}
    \STATE $O \gets \text{get\_accident\_info}()$
    \STATE $\mathbf{t}_n, D_n, R_n, S_n \gets \text{GPT}(P_r, O, \mathbf{Q})$ \hfill {\gray{\% Section~\ref{sec:reflection}}}
    \STATE $\mathbf{B} \gets \mathbf{B} \cup \{\mathbf{t}_n, D_n, R_n, S_n\}$ \hfill {\gray{\% update memory bank}}
\ENDIF
\end{algorithmic}
\end{algorithm}
\section{Related Works}

\subsection{Evolution of Vision Language Models}
Built on LLMs such as LLaMA~\cite{touvron2023llama, touvron2023llama2} and Vicuna~\cite{chiang2023vicuna}, a wide range of VLMs~\cite{alayrac2022flamingo, li2023blip, chen2022pali, li2023mimic, zhu2023minigpt, li2023videochat, liu2024visual, li2023monkey, zhang2023video, wang2023cogvlm, bai2023qwen, chen2024internvl} has emerged, extending their capabilities to multimodal understanding. These models have introduced innovative pretraining and fine-tuning techniques to bridge the gap between vision and language and raise extraordinary emergent abilities, typically including instruction following~\cite{zhang2023llava}, in-context learning~\cite{alayrac2022flamingo}, and chain-of-thought \cite{chen2024internvl} on multimodal data. 
Typically, VLMs utilize Q-former-like connectors~\cite{li2023blip2, bai2023qwen} or MLP-like connectors~\cite{zhang2023llava, chen2024internvl} for modal alignment and undergo three stages of training: pretraining, instruction-tuning, and optional alignment tuning to develop full-fledged capabilities.
For example, in addition to being pre-trained in large image-text corpora, models like LLaVA~\cite{liu2024visual} and MiniGPT-4~\cite{zhu2023minigpt} leverage instruction tuning to create versatile and interactive visual agents. Further advances include Qwen-VL~\cite{bai2023qwen}, which adopts a multistage training strategy to improve multilingual and fine-grained visual comprehension. 
InternVL~\cite{chen2024internvl} introduces a progressive alignment training strategy that facilitates the integrated understanding of multiple modalities, including text, images, videos, and others. These models have significantly broadened the scope of integration in vision languages, enabling more sophisticated applications.

\subsection{Leveraging Foundation Models for Autonomous Driving}
Following VLMs' great success, supervised fine-tuning and visual adaptation are conducted for various downstream vision tasks \cite{zhang2024vision}. In the realm of autonomous driving, numerous studies~\cite{fu2024drive, wen2023dilu, shao2023lmdrive, mao2023language, yuan2024rag, wen2024on} have explored the application of large foundation models, leveraging their embedded world knowledge and advanced reasoning capabilities. For example, DriveLM~\cite{sima2023drivelm} presents a graph visual question answering benchmark and incorporates VLMs to perform planning using a designed graph-based chain of thought, thereby enhancing the interpretability of autonomous driving systems. 
DriveMLM~\cite{wang2023drivemlm} utilizes VLMs to generate decision-making processes based on human instructions within simulated environments. 
ELM~\cite{zhou2024embodied} introduces a VLM tailored for embodied understanding in driving scenarios, while RAG-Driver~\cite{yuan2024rag} improves driving interpretation and signal prediction by combining retrieval-augmented generation (RAG) with in-context learning. Recent innovations, such as DriveVLM-Dual~\cite{tian2024drivevlm}, further integrate VLMs with data-driven planning pipelines, offering promising solutions for real-world deployment.

In contrast to the approaches mentioned above, we draw inspiration from human driving behaviors. Through the dual-process system, the memory bank, and the reflection mechanism, we implement the storage, updating, and transfer of experiences, enabling continuous exploration, learning, and improvement in closed-loop scenarios.

\vspace{-5pt}
\subsection{Transitioning from Data-Driven to Knowledge-Driven Autonomous Driving}
While data-driven approaches~\cite{yin2021center, li2022bevformer, liu2023bevfusion, jia2023think, shao2023safety, chitta2022transfuser, wu2022trajectory, hu2023planning, chitta2021neat, zhang2021end, codevilla2019exploring, chen2020learning, jiang2023vad} have achieved remarkable success in academia and industry, enabling the widespread adoption of autonomous driving technology in daily life, these methods often face inherent limitations. Specifically, their performance is tied to training data distribution, leading to adaptability issues and challenges in addressing long-tail scenarios when applied to diverse or unseen environments~\cite{peng2023tong, gildert2023building}. 
Besides, human drivers rely on deep common sense and a nuanced understanding of the world, allowing them to navigate unexpected scenarios effectively. This underscores the importance of transitioning toward knowledge-driven approaches that emphasize empirical reasoning, environmental induction, and the ability to develop specialized skills through continuous learning~\cite{liu2020overview, dou2023towards, zhang2023toward, xi2023rise}.

In the age of foundation models, LLMs and VLMs have shown exceptional abilities in reasoning and applying knowledge, proving effective in complex tasks like understanding, reasoning, and decision-making within autonomous driving~\cite{fu2024drive,liu2023can,cui2024survey}. Their extensive training on diverse datasets has equipped them with world knowledge and strong explanatory and reasoning skills, garnering significant interest from the research community~\cite{wayve2023lingo,ma2023dolphins,li2023towards,zhou2024embodied}.


\section{Methodology}
\begin{figure*}[t]
    \captionsetup{font={small}}
    \centering
    \includegraphics[width=1\linewidth]{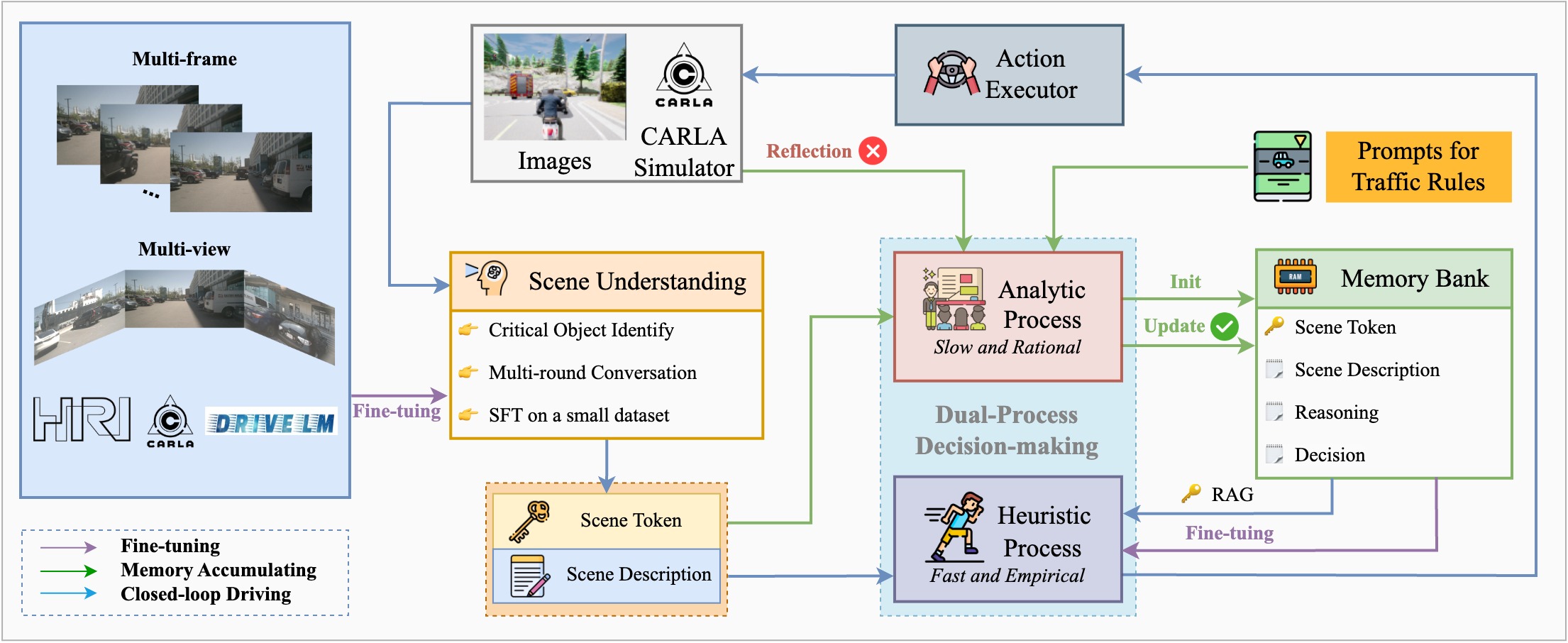}
      \caption{
      The architecture of {\newmethodname} consists of two primary modules: scene understanding and dual-process decision-making. The scene understanding module analyzes multi-view or multi-frame images, identifying critical objects and generating a scene token. This token serves as a characteristic representation of the current scene. The dual-process decision-making module then uses this scene description and the guidance of traffic rules to make reasoning and decisions. These decisions are converted into control signals to navigate the ego car in the simulator.
      Specifically, {\slowsystem} accumulates an initial memory bank used to train {\fastsystem} and updates it, especially when {\fastsystem} encounters accidents. {\fastsystem} leverages scene tokens to efficiently retrieve the most relevant historical scenarios from this memory bank, enabling rapid and informed driving decisions.
    }
    \label{fig:method}
\end{figure*}
\subsection{Overview}

Our proposed \newmethodname ~framework consists of four main components: the VLM for scene understanding~(Section~\ref{method:vlm}), Scene Encoder for extracting scene token~(Section~\ref{sec:scenetoken}), a dual-process decision-making module consisting of the {\slowsystem} (Section~\ref{method:slow}) and the {\fastsystem} (Section~\ref{method:fast}), which operates in conjunction with a controller detailed in Appendix \ref{sec:low-level-control}.
As illustrated in Algorithm~\ref{alg:alg1} and Figure~\ref{fig:method}, {\newmethodname} employs a VLM to analyze multi-frame images and describe key objects in the closed-loop simulator. The Scene Encoder creates scene tokens based on the current image and vehicle state. These object descriptions and scene tokens are passed to the Dual-Process Decision-Making Module, which performs scenario reasoning and determines driving actions. The resulting high-level decisions are then sent to the action actuator to generate control signals. Once an accident occurs, the system will automatically activate the reflection mechanism~(Section~\ref{method:slow}) for self-updating and continuous improvement.

\subsection{Scene Understanding with VLM} \label{method:vlm}
Human drivers focus on key objects around the vehicle to avoid information overload, improve reaction times, and reduce cognitive load. This strategy enhances concentration and lowers accident risks. Inspired by this, the scene understanding module in {\newmethodname} selectively identifies critical objects, simplifying the environment's description and easing the decision-making process.
\begin{figure*}[th]
    \centering
    \includegraphics[width=\linewidth]{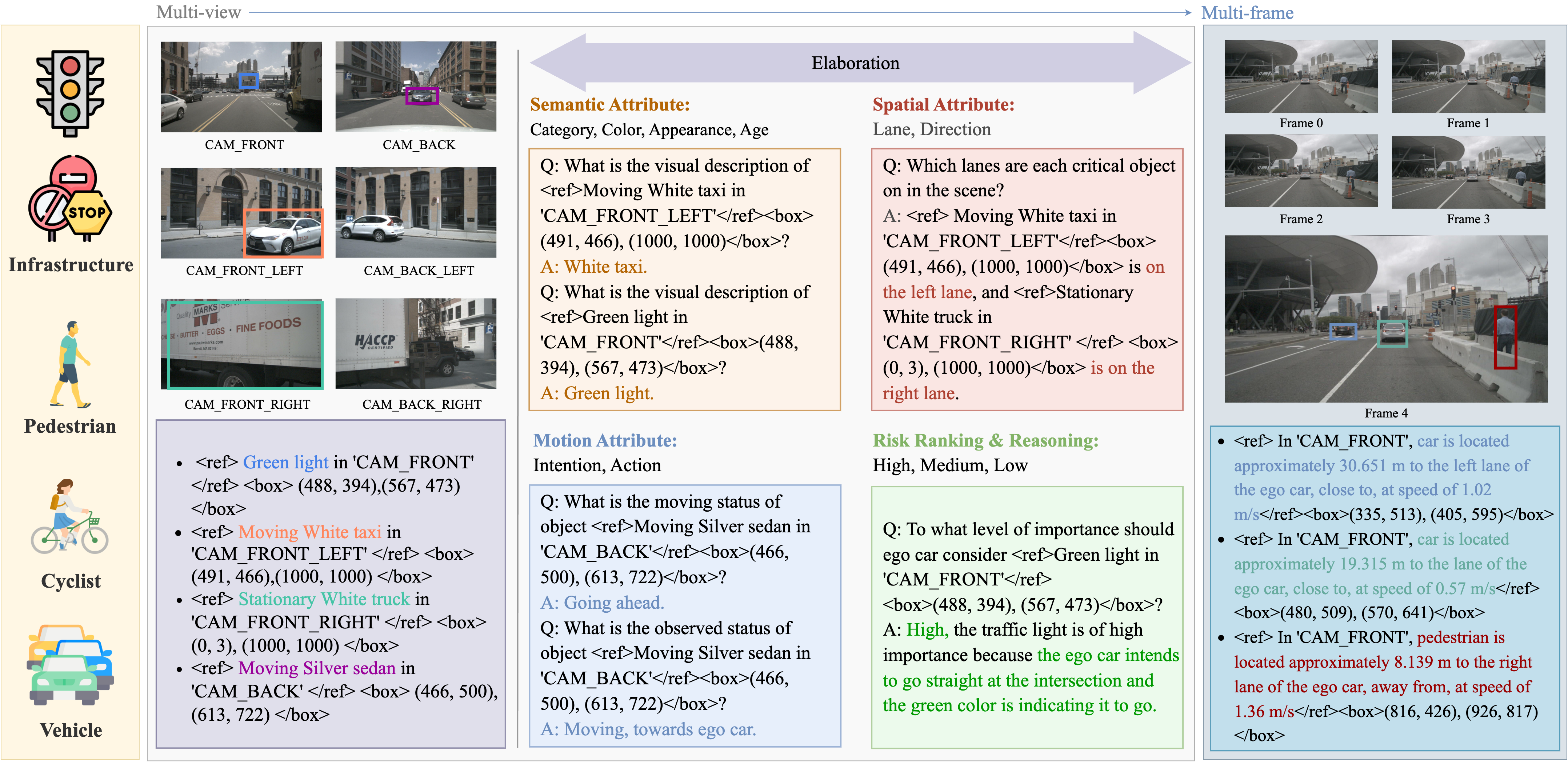}
    \caption{
    We create a dataset for instruction learning in VLM derived from DriveLM~\cite{sima2023drivelm}, Rank2Tell~\cite{sachdeva2024rank2tell}, and CARLA~\cite{dosovitskiy2017carla}. This dataset can be categorized into two types: multi-view and multi-frame. The multi-view annotations include a summary and elaboration, while the multi-frame annotations solely consist of a summary. Compared to multi-view annotations, the multi-frame annotations provide additional information such as exact velocity and motion trends.
    }
    \label{fig:data_anno}
\end{figure*}

Off-the-shelf foundation VLMs often lack domain-specific knowledge for driving. To address this, we perform supervised fine-tuning (SFT) and prompt the VLMs to generate detailed descriptions of objects that may impact driving decisions. These descriptions include semantic, spatial, and motion attributes and behavioral reasoning. By integrating these elements, the system achieves a more comprehensive understanding of the environment, improving safety and adaptability in complex driving scenarios.

To enhance the scene understanding capability of the VLM in autonomous driving, we develop a summary-elaboration data structure for SFT, as shown in Figure~\ref{fig:data_anno}. The data comprises two parts: multi-view and multi-frame. Notably, the summary is used only in the closed loop, while elaboration is reserved for training. For example, the descriptions generated by the VLM are represented as $\boldsymbol D=\{D_{s, i}, D_{l, i}, D_{m, i}, D_{r, i}\}_{i=0}^{N-1}$, where $N$ denotes the number of critical objects. In multi-view construction, each object includes:
\begin{enumerate}
    \item Semantic attribute $D_s$: Describes the object's category, typically important traffic participants (e.g., vehicles, cyclists), and infrastructure (e.g., traffic lights, stop signs).
    \item Spatial attribute $D_l$: Indicates its bounding box, lane position, and distance from the ego car, essential for safety and collision avoidance.
    \item Motion attribute $D_m$: Refers to the object's motion direction.
    \item Behavioral reasoning $D_r$: Explains the object's significance and influence on the ego car's driving decisions. For instance, a stop sign on the right is crucial when the ego car is going straight, as it indicates the need to stop at the intersection.
\end{enumerate}

For multi-frame data, dynamic attributes of objects within the scene are included in the summary, as illustrated in the right part of Figure~\ref{fig:data_anno}. Specifically, the VLM describes the location, motion trends, distance, and velocity of critical objects by analyzing the video data. Additionally, it provides the bounding boxes of these objects in the final frame.

\subsection{Scene Token} 
\label{sec:scenetoken}

To facilitate retrieving similar scenes in the memory bank for few-shot prompting, we propose a more efficient and precise method for extracting scene tokens. Our approach differs from {\methodname}~\cite{mei2024continuously} in that we can avoid the ambiguity, synonyms, and paraphrasing of scene descriptions, resulting in improved performance compared to text embeddings.
In this paper, {\newmethodname} generates meta-actions for steering and speed regulation, mirroring the primary control mechanisms employed by human drivers. Our key insight is that scenes requiring similar human control responses (regarding steering and braking) can be considered analogous at the control level. Building upon this observation and inspired by ACO \cite{zhang2022learning}, we develop a comparative learning approach that operates in two distinct spaces: the Action space (ACT) for steering and the Acceleration space (ACC) for braking. This dual-space comparative learning framework enables us to derive scene tokens that capture the underlying control similarities.

\subsubsection{Scene Encoder} 
\begin{figure}[t]
    \captionsetup{font={small}}
    \centering
    \includegraphics[width=\linewidth]{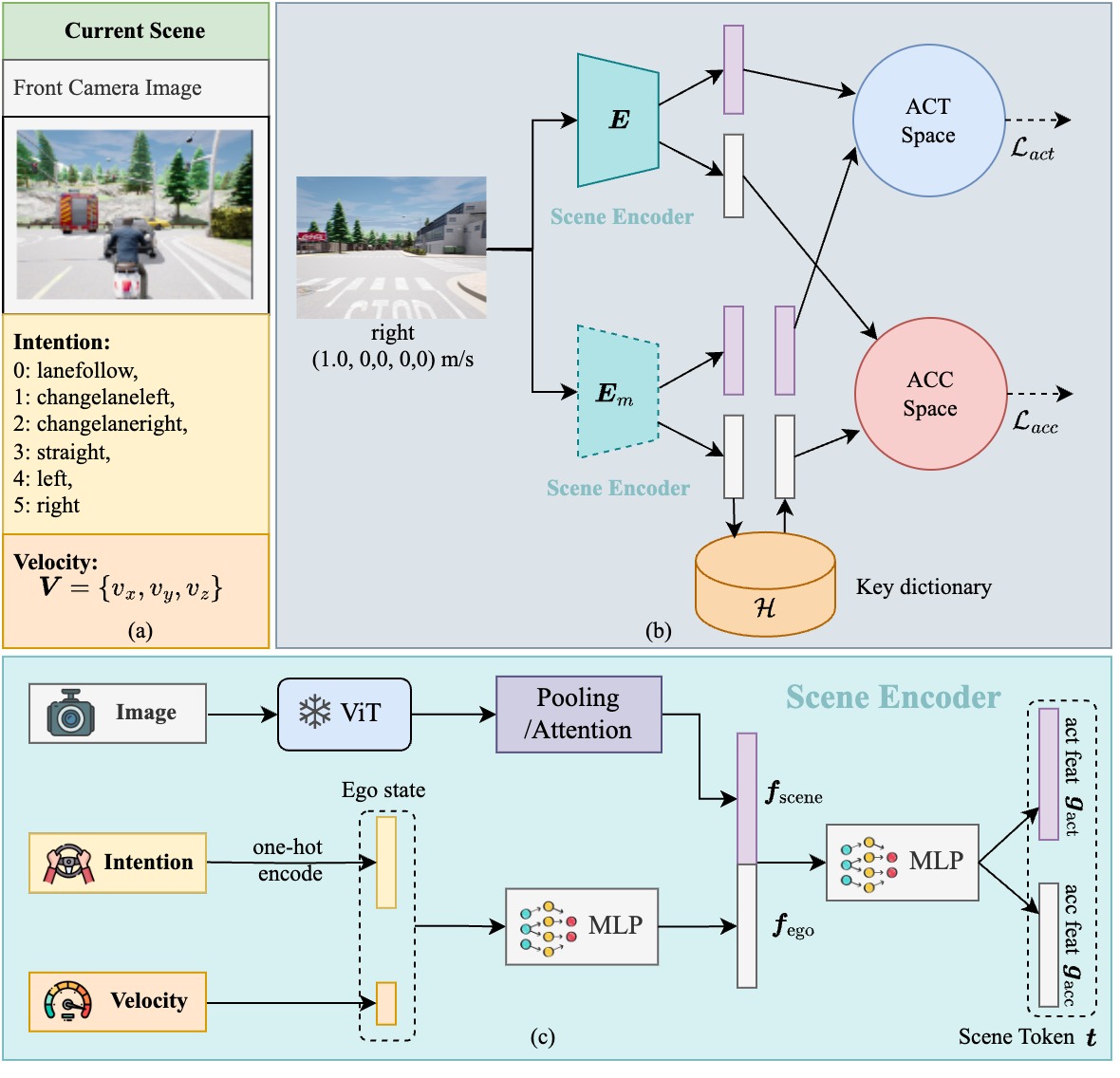}
    \caption{
    The training pipeline of our Scene Encoder is outlined as follows: (a) provides details about the input data; (b) illustrates how we form both ACT and ACC for the input images and update the model using contrastive loss in these two spaces; (c) presents the architecture of the Scene Encoder.
    }
    \label{fig:scenetoken}
    \vspace{0pt}
\end{figure}

As illustrated in Fig.~\ref{fig:scenetoken}, given a batch of image $\boldsymbol{I} \in \mathbb{R}^{B\times H \times W \times 3}$ and the ego state $\boldsymbol{A}\in \mathbb{R}^{B\times 9}$, we process the input through the Scene Encoder $\boldsymbol E$ to derive the Scene Token $\boldsymbol{t}\in \mathbb{R}^{B\times 256} $, which encompasses $\boldsymbol{g}_{\text{act}}\in \mathbb{R}^{B\times 128}$ and $\boldsymbol{g}_{\text{acc}}\in \mathbb{R}^{B\times 128}$. 
A momentum-updated Scene Encoder $\boldsymbol E_m$ is used to extract feature vector $\boldsymbol{t}' \in \mathbb{R}^{B\times256}$. The updated rule of the encoder's parameters is:
\begin{align}
    \theta_{\boldsymbol{E_m}} \leftarrow \alpha \theta_{\boldsymbol{E_m}} + (1-\alpha)\theta_{\boldsymbol{E}}
\end{align}
where $\alpha$ is the momentum coefficient and $\theta_{\boldsymbol{E_m}}, \theta_{\boldsymbol{E}}$ are the parameters of $\boldsymbol{E}_m$ and $\boldsymbol{E}$. 

Specifically, the $\text{intent}$ is derived through one-hot encoding. The features are concatenated with ego velocity $\boldsymbol{V}$ via an MLP to generate the ego state features $\boldsymbol{f}_{\text{ego}} \in \mathbb{R}^{B\times 256}$. The image is processed to extract ViT features $\boldsymbol{f}_{\text{img}} \in \mathbb{R}^{B\times N\times C}$, which are then compressed into the scene features $\boldsymbol{f}_{\text{scene}} \in \mathbb{R}^{B\times 256}$ using either max-pooling or attention mechanisms. The final Scene Token $\boldsymbol{t}$ is produced by concatenating $\boldsymbol{f}_{\text{ego}}$ and $\boldsymbol{f}_{\text{scene}}$ after applying the MLP.

\subsubsection{Key Dictionary} 
Following MoCo~\cite{he2020momentum}, we use a key dictionary $\mathcal{H}$ to store historical encoded Scene Token $\boldsymbol{t}'$ to enable larger contrastive batch size.
In this study, we utilize the Scene Encoder $\boldsymbol E_m$ to generate the Scene Token $\boldsymbol t'$. This process establishes positive pairs with the current training features. It ensures that these tokens are stored in the memory bank, $\mathcal H$, to form positive and negative sample pairs in subsequent training iterations. When the size of $\mathcal H$ exceeds its predetermined capacity, the samples within the dictionary are systematically replaced.
\subsubsection{Traning Loss}
During training, a batch of images and their corresponding ego states are sampled. Following data augments, these images are encoded by both $\boldsymbol{E}$ and $\boldsymbol{E_m}$ to obtain the query features $\boldsymbol{g}_{\rm {sp}} \in \mathbb{R}^{B\times 128}$ and key features $\boldsymbol g'_{\rm {sp}}\in \mathbb{R}^{B\times 128}$. Simultaneously, $N$ key features are sampled from both $\mathcal{H}$ and $\boldsymbol g'_{\rm {sp}}$ to form the key set $\boldsymbol{K}$. The losses $\mathcal{L}_{\text{act}}$ and $\mathcal{L}_{\text{acc}}$ are then computed as:
\begin{align}
    \mathcal L_{\text{sp}} = - \text{log} \dfrac{\sum_{z^+ \in P_{\text{sp}}(\boldsymbol g_{\text{sp}}) \text{exp}}(\boldsymbol{g}_{\text{sp}}\cdot z^{+}/\tau)}{\sum_{z^- \in N_{\text{sp}}(\boldsymbol g_{\text{sp}}) \text{exp}}(\boldsymbol{g}_{\text{sp}}\cdot z^{-}/\tau)}
\end{align}
where $P_{\text{sp}}(\boldsymbol t_{\text{sp}})$ and $N_{\text{sp}}(\boldsymbol{t}_\text{sp})$ are the  positive and negative keys in ``$\text{sp}$'' space respectively. They are represented as:
\begin{align}
\begin{array}{rl}
P_{\text{sp}}(\boldsymbol g_{\text{sp}}) &= \{z\, | \,\Vert \hat{a}'_\text{sp}  - \hat{a}_\text{sp}\Vert < \sigma_{\text{sp}}, (z, \hat{a}'_\text{sp}) \in \boldsymbol{K} \}, \\    N_{\text{sp}}(\boldsymbol{g}_\text{sp}) &  \equiv \boldsymbol{K} \setminus P_{\text{sp}}(\boldsymbol g_{\text{sp}}) , \\
    \text {sp} &\in \{\text{act}, \text{acc}\},
\end{array}
\label{fun:positiveandnegative}
\end{align}
 where $\hat{a}_\text{sp}$ is the label of the query feature, and $\hat{a}'_\text{sp}$ is the label of the key feature, with the values of steering and brakes in the ACT and ACC spaces, respectively. Also, $\sigma_{\text{sp}}$ is the distance threshold in ``$\text{sp}$'' space. In the ACC space, a negative pair is identified when one vehicle brakes and the other does not.

The overall contrastive loss of Scene Token is:
\begin{align}
    \mathcal{L} = \lambda_{\text{act}}\mathcal{L}_{\text{act}} + \lambda_{\text{acc}}\mathcal{L}_{\text{acc}}
\end{align}
where $\lambda_{\text{act}}$ and $\lambda_{\text{acc}}$ are hyper-parameters used to adjust the weights of $\mathcal{L}_{\text{act}}$ and $\mathcal{L}_{\text{acc}}$.

\subsection{\slowsystem} \label{method:slow}
Leveraging scene descriptions offered by the VLM, we develop the {\slowsystem}, a framework designed to emulate the logical reasoning processes characteristic of human drivers. The {\slowsystem} utilizes logical reasoning to navigate complex scenarios, employing structured analysis and rational decision-making to ensure safety in driving tasks.

Through extensive pre-training on diverse datasets, LLMs inherently accumulate a vast repository of world knowledge, equipping them to address complex problems with nuanced reasoning and understanding~\cite{wen2023dilu}. This capability meets the demand of the {\slowsystem}, which relies on thorough analysis and contextual awareness to make informed decisions in driving scenarios. Our {\slowsystem} leverages the world knowledge embedded in LLMs to interpret scene descriptions and produce high-quality driving decisions. Empirical results indicate that incorporating specific traffic rules, as outlined in Appendix~\ref{sec:prompt}, further enhances the system's safety and reliability in real-world driving scenarios.
Moreover, we combine the VLM with the {\slowsystem} to conduct closed-loop experiments, enabling the collection of high-quality decision-making data and outcomes. These results, stored as ``experience'' in a memory bank, are incrementally accumulated and can be effectively transferred to the {\fastsystem}. This transfer empowers the {\fastsystem} to leverage prior experience for rapid response in analogous scenarios, as elaborated in Section \ref{method:fast}.

\textit{Reflection mechanism:} 
The {\slowsystem} is employed to facilitate reflection on traffic accidents, as demonstrated at the bottom of Algorithm~\ref{alg:alg1}. Precisely, in a closed-loop driving scenario integrating the VLM and {\fastsystem}, the occurrence of any accident $O$ activates a reflective mechanism. The {\slowsystem} subsequently analyzes the scene description $\boldsymbol D$, reasoning $\boldsymbol R$, and decision $\boldsymbol S$ from the historical frames $\boldsymbol{Q}$ preceding the incident. This analysis identifies causal factors, detects errors, and proposes corrective reasoning $R_n$ alongside decision-making strategies $S_n$. A detailed example of the reflection process is provided in Appendix~\ref{sec:reflection}.

The insights derived from this reflection procedure are subsequently integrated into the system’s memory bank, 
enabling {\newmethodname} to learn continuously from past failures and make progressively better-informed decisions in future driving scenarios. Notably, the accumulated experience in the memory bank exhibits strong transferability and generalization, allowing it to be directly applied to other lightweight models and easily adapted to diverse scenarios, as discussed in Section \ref{ab:generalization}.

\subsection{\fastsystem} \label{method:fast}
\subsubsection{LLM}
Although the {\slowsystem} excels at providing precise driving reasoning and decisions through its detailed analysis and thorough evaluation, it is inherently slow processing often results in duplicated and redundant efforts, limiting its practicality in real-world driving scenarios. 
Drawing inspiration from human driving behavior, where drivers develop muscle memory through repeated practice that enables efficient reactions with minimal cognitive load, we introduce a {\fastsystem} within {\newmethodname} incorporating a lightweight language model.

To enable effective knowledge transfer, we apply supervised fine-tuning (SFT) using the data accumulated in the memory bank, as described in Section \ref{method:slow}. This process distills knowledge from the {\slowsystem} into the lightweight model, allowing the {\fastsystem} to adapt its behavior to diverse scenarios while operating significantly faster {(approximately five times faster in our experiments)}. Our previous results~\cite{mei2024continuously} reveal that the lightweight model without SFT fails to produce reliable driving decisions.

\subsubsection{few-shot prompting}
We utilize few-shot prompting~\cite{wen2023dilu,mei2024continuously} to enhance the {\fastsystem}'s generalization for unseen scenes and reduce hallucinations, leading to more robust decisions. This approach allows the {\fastsystem} to effectively draw on insights from its memory bank, improving the accuracy of future driving decisions.

\begin{figure}[t]
    \captionsetup{font={small}}
    \centering
    \includegraphics[width=\linewidth]{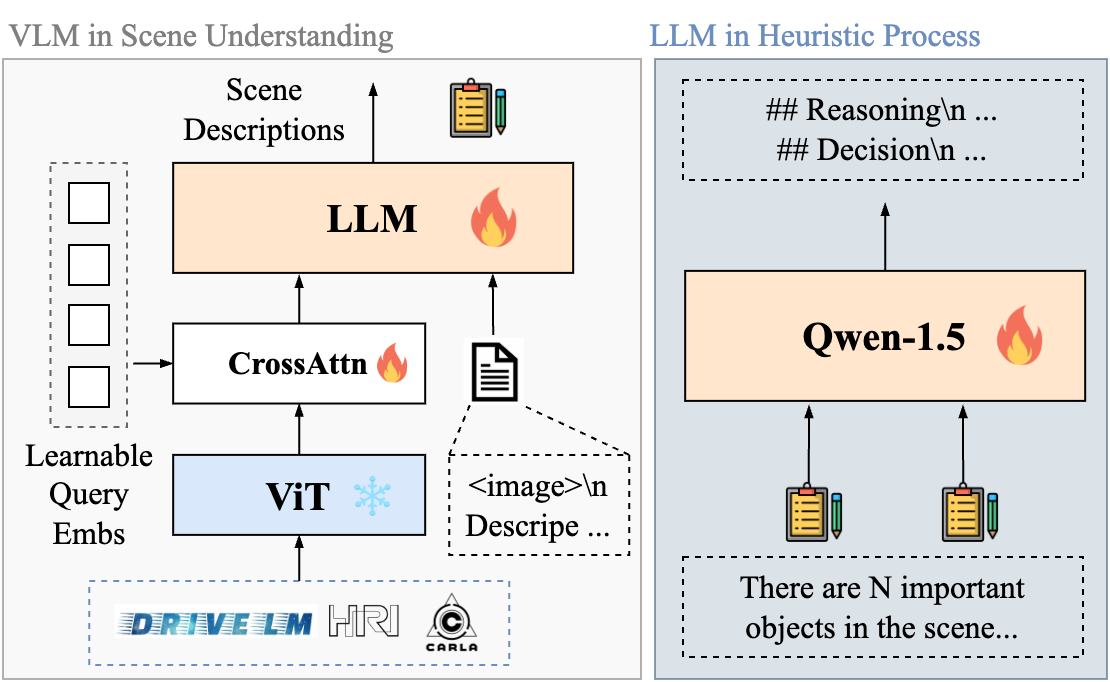}
    \caption{
    The illustration depicts the fine-tuning process. Figure (a) illustrates the fine-tuning of the VLM using 4.1K instruction-following data points for scene understanding. Figure (b) shows the utilization of the collected samples in the memory bank to fine-tune Qwen-1.5, employed in the {\fastsystem} model.
    }
    \label{fig:finetune}
    \vspace{-5pt}
\end{figure}
Following Section~\ref{sec:scenetoken}, we can obtain the scene token $\boldsymbol{t}_{\rm q}$ for the current scene. Subsequently, the cosine similarity between $\boldsymbol{t}_{\rm q}$ and the scene tokens $\{\boldsymbol t_i\}_{i=0}^{M-1}$ in the memory bank with the size of $M$ is calculated by:
\begin{equation}
    \rm{cosine}(\boldsymbol{t}_{\rm q}, \boldsymbol{t}_{i}) = \frac{\boldsymbol{t}_{\rm q}\cdot\boldsymbol{t}_{i}}{\|\boldsymbol{t}_{\rm q}\| \|\boldsymbol{t}_{i}\|}.
    \label{fun:similarity}
\end{equation}

We select the top-k samples with the highest similarity scores as queried scenes. The scene descriptions $\{D_i\}_{i=0}^{k-1}$, reasoning $\{R_i\}_{i=0}^{k-1}$, and decisions $\{S_i\}_{i=0}^{k-1}$ of these samples, along with the current scene description ${D}_c$, are input into the {\fastsystem} for final reasoning ${R}_c$ and decision ${S}_c$.
\section{Experiments}

\subsection{Data preparation}
\label{sec: exp-data}
\subsubsection{Data for VLM} 
\label{sec:exp-data-vlm}
We constructed an instruction-following dataset for the supervised fine-tuning of VLM by integrating data collected from Rank2Tell~\cite{sachdeva2024rank2tell}, DriveLM~\cite{sima2023drivelm}, and CARLA~\cite{dosovitskiy2017carla}. We standardized the referring format for key objects as follows: \texttt{<ref>In {camera view}, {properties}</ref><box>{coordinates}</box>}. Each dataset was processed with the original labels to obtain unique question-answer pairs. Overall, the dialogue is constructed summary-elaborated, as shown in Figure~{\ref{fig:data_anno}}. The elaboration includes four aspects of key object properties: semantic attributes, spatial attributes, motion attributes, and importance. Apart from multi-view dialogues, we added summaries for multi-frame images, introducing properties such as distance, speed, and motion trends of key objects. We constructed dialogues using five frames, with Rank2Tell operating at 10 Hz and the others at 2 Hz. We collected $5\rm{K}$ of multi-view summary data and $2\rm{K}$ of multi-frame summary data from CARLA Towns 01-04, 06, 07, and 10 to train the VLM.

\subsubsection{Data for {\fastsystem}}
\label{sec:dataforfast}
We accumulate experience in a closed-loop setting and store it in a memory repository by integrating {\slowsystem} and VLM, which serves as few-shot examples for subsequent SFT and {\fastsystem} prompting. Our approach also incorporates a dynamic updating mechanism to address issues encountered by {\fastsystem}, as detailed in Section \ref{method:slow}. It is worth noting that these samples are collected in a closed-loop environment without human intervention. 
As with {\methodname}~\cite{mei2024continuously}, we used a memory bank size of 18.1k in Table~\ref{tab:carla-leaderboard} and used it as the default in this paper.

\subsubsection{Data for Scene Encoder}
We utilized data collected from CARLA and the nuScenes~\cite{caesar2020nuscenes} (train set) to train the Scene Encoder. Specifically, we collected 90K driving frames from the CARLA simulator, of which 70\% were designated for training.  Each frame includes an image captured by the front-view camera and data on intent, velocity, steering, and braking. Steering values are normalized within the [-1, 1] range, whereas braking values range from [0, 1]. We use steering and braking as labels for the ACT and ACC spaces, respectively, and set the distance threshold $\sigma_{\text{sp}}$ ($\text{sp} \in \{\text{act}, \text{acc}\}$) to 0.04.

\begin{table*}[t]
    \centering
    \small
    \renewcommand{\arraystretch}{1.1}
    \caption{Comparison of our {\newmethodname} with competitive methods on Town05 Short benchmark. 
    Notably, {\newmethodname} achieved superior performance while utilizing less data. In this context, ``L'' and ``C'' represent LiDAR and camera modalities. Additionally, ``DD'' and ``KD'' abbreviate data-driven and knowledge-driven approaches, while ``OE'' and ``ST'' correspond to OpenAI embedding and scene token, respectively. The method marked with ``$^\dagger$'' utilizes InternVL2-8B~\cite{chen2024internvl} as the VLM, and the method indicated by ``$^*$'' employs {\slowsystem}.}
    \resizebox{0.85\linewidth}{!}{
    \begin{tabular}{l|c|ccc|cc|cc}
    \toprule
        Row & {Method} & {Modality} & Type & Annotations & Retrieval& Reflection  & {DS $\uparrow$} & {RC $\uparrow$} \\ 
        \midrule
        01 &InterFuser~\cite{shao2023safety} & L+C & DD & $3\rm{M}$& -& -  & \textbf{94.95$\pm$1.91} & \textbf{95.19$\pm$2.57} \\ 
        02 &TransFuser~\cite{chitta2022transfuser} & L+C & DD  & $ 228\rm{K}$ & - & - & 54.52$\pm$4.29 & 78.41$\pm$3.75 \\
        \midrule
        03 &VAD~\cite{jiang2023vad} & C & DD & 228\rm{K} & - & - & 64.30 & 87.30 \\
        04 &NEAT~\cite{chitta2021neat} & C & DD & $130\rm{K}$ & -& -  & 58.70$\pm$4.11 & 77.32$\pm$4.91 \\
        05 &Roach~\cite{zhang2021end} & C & DD & - & - & - & 65.26$\pm$3.63 & 88.24$\pm$5.16 \\
        06 &WOR~\cite{chen2021learning} & C & DD & $1\rm{M}$ & -& -  & 64.79$\pm$5.53 & 87.47$\pm$4.68 \\
        07 &LBC~\cite{chen2020learning} & C & DD & $157\rm{K}$ & - & - & 30.97$\pm$4.17 & 55.01$\pm$5.14 \\
        08 &CILRS \cite{codevilla2019exploring} & C & DD & $720\rm{K}$ & -& -  & 7.47$\pm$2.51 & 13.40$\pm$1.09  \\ 
        09 &{{\methodname}}$^*$~\cite{mei2024continuously} & C & KD & $11\rm{K}$ & -& -  & 81.31$\pm$2.37 & 94.22$\pm$3.18 \\ 
        10 &{{\methodname}}~\cite{mei2024continuously} & C & KD & $11\rm{K}$ & OE & $\times$  & 75.73$\pm$1.36 & 92.10$\pm$1.44 \\
        11 &{{\methodname}}~\cite{mei2024continuously} & C & KD  & $11\rm{K}$ & OE & \checkmark  & {83.11$\pm$0.28} & {94.98$\pm$0.54} \\
        \rowcolor[HTML]{ECECEC} 12 &\textbf{{\newmethodname}$^*$} & C & KD & $41\rm{K}$ & -& $\times$  & 86.55$\pm$3.12 & 97.19$\pm$0.42 \\
        \rowcolor[HTML]{ECECEC} 13 &\textbf{{\newmethodname}} & C & KD & $41\rm{K}$ & OE & $\times$  &{82.58$\pm$3.04} & {96.68$\pm$1.54} \\
        \rowcolor[HTML]{ECECEC} 14 &\textbf{{\newmethodname}} & C  & KD & $41\rm{K}$ & ST & $\times$  &{83.78$\pm$2.01} & {99.42$\pm$0.03} \\
        \rowcolor[HTML]{ECECEC} 15 &\textbf{{\newmethodname}} & C & KD & $41\rm{K}$ & ST& $\checkmark$  &\textbf{88.19$\pm$2.98} & \textbf{99.53$\pm$0.17} \\
        \rowcolor[HTML]{ECECEC} 16 &\textbf{{\newmethodname}}$^\dagger$ & C & KD & $41\rm{K}$ & ST& $\checkmark$  &{80.26$\pm$2.46} & {98.80$\pm$0.67} \\ 
        \bottomrule
    \end{tabular}
    }
    \label{tab:carla-leaderboard}
\end{table*}

\subsection{{Implementation Details}}
\label{sec: exp-detail}
According to Section~\ref{sec:vlm}, we utilize Qwen-VL-7B~\cite{bai2023qwen} and InternVL2-8B~\cite{chen2024internvl} as the visual language model (VLM) for scene understanding, employ GPT-4o as the slow system (\slowsystem) for reasoning and logic, and use Qwen1.5-1.8B~\cite{qwen} as the fast system (\fastsystem) for automatic and rapid thinking in {\newmethodname}. The Scene Encoder $\boldsymbol{E}$ implements ``pooling + state'' to extract the scene token, where the input includes the ego state, and max pooling is used to extract $\boldsymbol{f}_{\rm{scene}}$, as demonstrated in Table \ref{table:precision@1}.

To fully exploit the capabilities of our VLM in closed-loop driving experiments, we conducted SFT with instruction-based data as discussed in Section~\ref{sec:exp-data-vlm}. For Qwen-VL-7B~\cite{bai2023qwen}, we utilized the AdamW optimizer \cite{loshchilov2017decoupled} with hyperparameters $\beta_1 = 0.9$ and $\beta_2 = 0.95$. We combined this with a cosine decay of the learning rate, initially set to $1 \times 10^{-5}$. The batch size was set to 8, and the model was trained for 5 epochs on 8 A100 GPUs, taking approximately 69 hours. For InternVL2-8B~\cite{chen2024internvl}, we employed the same optimizer with hyperparameters $\beta_1 = 0.9$ and $\beta_2 = 0.99$. The initial learning rate was set to $4 \times 10^{-5}$ and decayed using a cosine schedule. Similarly, the batch size was 8, and the model was trained for 5 epochs on 8 A100 GPUs, requiring around 7 hours. The input image resolution was set at $448 \times 448$ pixels.

For {\fastsystem}, we conducted SFT on Qwen1.5-1.8B for 5 epochs using samples stored in the memory bank, taking about 6 hours. The training hyperparameters are consistent with VLM's training procedure. Figure \ref{fig:finetune} shows the detailed fine-tuning process. The dual-process decision module outputs meta-actions (e.g., ``AC'', ``DC'', ``LCL'', ``LCR'', ``IDLE'', ``STOP'') at a frequency of 2 Hz, which are further refined into control signals, as detailed in Appendix~\ref{sec:low-level-control}. 

For the scene token, we trained the Scene Encoder on our custom training set, consisting of 63K CARLA data samples and the nuScenes dataset, for 12 epochs, taking approximately one hour. We utilized a batch size of 128 and the SGD optimizer with hyperparameters set as follows: $\lambda_{\rm{act}} =\lambda_{\rm{acc}}= 1$, a learning rate of $0.03$, a weight decay of 1e-4, and an $\mathcal H$ size of 4096. For image augmentation, we employed DriveArena~\cite{yang2024drivearena} to generate two batches with aspect ratios of $336\times600$ and $448\times800$, respectively, while ensuring consistency with the nuScene dataset in terms of layouts and boxes. In Figure~\ref{fig:scenetoken}, ``pooling'' refers to the application of max pooling to generate the scene feature $\boldsymbol{f}_{\rm{scene}}$ from ViT features, whereas ``attention'' denotes the use of a learnable weight matrix to create the scene feature $\boldsymbol{f}_{\rm{scene}}$ from ViT features.

\begin{table}[htbp]
    \centering
    \small
          \setlength{\tabcolsep}{1.1pt}
    \renewcommand{\arraystretch}{1.1}
    \caption{Comparison of our {\newmethodname} with competitive methods in Town05 Long benchmark.}
    \resizebox{\linewidth}{!}{
    \begin{tabular}{l|ccc|ccc}
		\toprule
        {Method} & {Modality} & Type & Annotations & {DS (\%) $\uparrow$} & {RC (\%) $\uparrow$} & {IS (\%) $\uparrow$} \\ 
        \midrule
        DriveMLM~\cite{wang2023drivemlm} & L+C & DD & $ 2\rm{M}$ & \textbf{76.1} & \textbf{98.1} & \textbf{78.0 }\\
        ThinkTwice~\cite{jia2023think} & L+C & DD & $2\rm{M}$ & 70.9 & 95.5 & 75.0 \\ 
        InterFuser~\cite{shao2023safety} & L+C & DD & $3\rm{M}$ & 68.3 & 95.0 & 72.0 \\ 
        TransFuser~\cite{chitta2022transfuser} & L+C & DD & $ 228\rm{K}$ & 31.0 & 47.5 & 77.0 \\
        \midrule
        VAD~\cite{jiang2023vad} & C & DD & 228$\rm{K}$ & 30.3 & 75.2 & - \\
        TCP~\cite{wu2022trajectory} & C & DD & $420\rm{K}$ & 57.2 & 80.4 & 73.0 \\
        NEAT~\cite{chitta2021neat} & C & DD & $130\rm{K}$ & 37.7 & 62.1 & 61.0 \\
        Roach~\cite{zhang2021end} & C & DD & - & 43.6 & 80.4 & 54.4  \\
        WOR~\cite{chen2021learning} & C & DD & $1\rm{M}$ & 44.8 & 82.4 & 54.0 \\
        LBC~\cite{chen2020learning} & C & DD & $157\rm{K}$ & 7.1 & 32.1 & 22.1 \\
        CILRS \cite{codevilla2019exploring} & C & DD & $720\rm{K}$ & 3.7 & 7.2 & 51.4  \\ \hline
        \textbf{\methodname}~\cite{mei2024continuously} & C & KD & $11\rm{K}$ & 51.7 & \textbf{100} & 51.7 \\
        \rowcolor[HTML]{ECECEC}\textbf{\newmethodname} & C & KD & $41\rm{K}$ & \textbf{73.7} & {95.7} & \textbf{78.0}\\
        \bottomrule
    \end{tabular}
    }
    \label{tab:carla-long-leaderboard}
\end{table}

\subsection{Evaluation in Closed-Loop Driving}
\label{exp: close-loop}
\subsubsection{CARLA Leaderbord}
\label{exp: close-loop-carla}
We conduct closed-loop experiments in CARLA, a popular open-source simulator, to evaluate the performance of our {\newmethodname}. 
To assess our method's effectiveness, we perform a detailed evaluation in a closed-loop driving environment using the Town05 benchmark. We utilize several metrics: Driving Score (DS), Route Completion (RC), and Infraction Score (IS). RC represents the percentage of the route completed by the agent, while IS reflects the penalties from accidents. The final metric, DS, is calculated by combining RC and IS, allowing us to evaluate the driving performance on a given route.

Specifically, we provide five different configurations to thoroughly evaluate our method, as shown in the highlighted rows in Table I. From top to bottom, these configurations are: \emph{\romannumeral 1)} closed-loop experiments using the {\slowsystem} directly; \emph{\romannumeral 2)} {\newmethodname} with OpenAI Embedding as the scene token and a memory bank without reflection; \emph{\romannumeral 3)} using our Scene Encoder to extract the scene token and a memory bank without reflection; \emph{\romannumeral 4)} employing our Scene Encoder for scene token extraction and a memory bank with reflection; and \emph{\romannumeral 5)}, which utilizes InternVL2-8B~\cite{chen2024internvl} as the VLM, while the rest remains consistent with \emph{\romannumeral 4)}. For additional VLM ablation studies, refer to Section~\ref{sec:vlm}.

As shown in Table~\ref{tab:carla-leaderboard}, our {\newmethodname} outperforms all camera-only methods, including the previous {\methodname} approach. Notably, our technique outperforms even TransFuser \cite{chitta2022transfuser}, which utilizes LiDAR sensor input. 
While our method shows slightly lower performance compared to InterFuser~\cite{shao2023safety}, another multi-sensor approach, it is noteworthy that InterFuser requires a substantially larger dataset with approximately 73 times more camera and LiDAR annotations.
The comparison between entries in rows 09 and 12, as well as rows 10 and 13, shows the performance improvement of our proposed temporal module over LeapAD~\cite{mei2024continuously}, fully confirming the superiority of employing multi-frame data for analysis. Replacing the previous OpenAI Embedding with our designed Scene Encoder (as seen in rows 13 and 14) yields notable improvements in DS and RC, providing empirical support for the Scene Encoder's effectiveness. We further substantiate the advancements of the Scene Encoder in Section~\ref{sec:scenetoken} of the ablation experiment. 
Finally, the comparison between rows 15 and 12 highlights the efficacy of our dual-process decision module, with {\newmethodname} achieving superior DS and RC scores.

In addition, we present the evaluation results on the more challenging Town05 Long benchmark in Table~\ref{tab:carla-long-leaderboard}. The results indicate that {\newmethodname} significantly improves over the previous {\methodname} and outperforms all other vision-only methods. Notably, {\newmethodname}'s performance is comparable to that of DriveMLM, despite the latter being trained with LiDAR and 48 times more data.

\subsubsection{DriveArena}
\label{exp: close-loop-arena}

\begin{table*}[t]
    \centering
    \setlength{\tabcolsep}{1.1pt}
    \renewcommand{\arraystretch}{1.1}
    \caption{Comparison of our {\newmethodname} with end-to-end autonomous driving algorithms in DriveArena\cite{yang2024drivearena}. We evaluated the {\newmethodname} with InternVL2 and QwenVL for scene understanding. To demonstrate the effectiveness of the memory bank, we also tested the {\newmethodname} with and without the memory bank module in DriveArena\cite{yang2024drivearena}. The calculated PDMS, RC, and ADS are presented in percentages.}
     \resizebox{\linewidth}{!}{
     \begin{tabular}{l|ccc|ccc|ccc|ccc|ccc}
    \Xhline{0.5pt}
        \multirow{2}{*}{\textbf{Method}} & \multicolumn{3}{c|}{\textbf{DA sing\_route\_1}} & \multicolumn{3}{c|}{\textbf{DA sing\_route\_2}} & \multicolumn{3}{c|}{\textbf{DA boston\_route\_1}} & \multicolumn{3}{c|}{\textbf{DA boston\_route\_2}} & \multicolumn{3}{c}{\textbf{Avg.}} \\
        &{PDMS$_\uparrow$} & {RC$_\uparrow$} & {ADS$_\uparrow$} & {PDMS$_\uparrow$} & {RC$_\uparrow$} & {ADS$_\uparrow$} &{PDMS$_\uparrow$} & {RC$_\uparrow$} & {ADS$_\uparrow$} & {PDMS$_\uparrow$} & {RC$_\uparrow$} & {ADS$_\uparrow$} & {PDMS$_\uparrow$} & {RC$_\uparrow$} & {ADS$_\uparrow$}\\ 
        \hline

        VAD & 53.15 & 4.67 & 2.48 
                         & 51.47 & 4.00 & 2.06 
                         & 58.30 & 6.04 & 3.52 
                         & 50.54 & 3.66 & 1.85 & 53.36±3.46 & 4.59±1.05 & 2.48±0.74\\ 

        UniAD & 76.15 & 16.84 & 12.82 
                         & 72.15 & 16.9 & 8.75 
                         & 49.52 & 9.1 & 4.50 
                         & 68.88 & 12.1 & 8.35 & 66.68±11.82 & 13.74±3.82 & 8.60±3.40\\ 
        \hline
        \rowcolor[HTML]{ECECEC}{\newmethodname} + InternVL2 (Mem.) & 78.92 & \textbf{98.52} & 77.75
                                  & 90.67 & \textbf{56.79} & 51.49
                                  & 72.39 & 14.36 & 10.40
                                  & 87.92 & 44.00 & 38.69 & 82.48±8.39 & \textbf{53.42±34.93} & 44.58±27.99\\ 
        
        \rowcolor[HTML]{ECECEC}{\newmethodname} + QwenVL & 81.03 & 98.51 & 79.82 
                         & 94.23 & 51.44 & 48.47 
                         & 82.53 & 18.51 & 15.28 
                         & 84.48 & 12.08 & 10.21
                         & 85.57±5.95 & 45.14±39.54 & 38.45±32.38\\ 
        
        \rowcolor[HTML]{ECECEC}{\newmethodname} + QwenVL (Mem.) & \textbf{82.33} & 98.44 & \textbf{81.05} 
                                 & \textbf{95.70} & 54.42 & \textbf{52.08}
                                 & \textbf{85.41} & \textbf{18.66} & \textbf{15.94}
                                 & \textbf{93.19 }& \textbf{35.42} & \textbf{33.01}
                                 & \textbf{89.16±6.31} & 51.74±34.39 & \textbf{45.52±27.90}\\
        
        \Xhline{0.5pt}

    \end{tabular}
    }
    \label{tab:drivearena}
\end{table*}

We conducted closed-loop experiments in DriveArena\cite{yang2024drivearena}, a high-fidelity closed-loop simulation platform, to evaluate {\newmethodname} 's driving performance in a simulation environment resembling real-world conditions. These experiments were performed on four pre-defined routes, with two selected in Boston and two in Singapore. The evaluation utilized three key metrics: PDM Score (PDMS), Route Completion (RC), and Arena Driving Score (ADS). PDMS, initially introduced by NAVSIM\cite{dauner2024navsim}, evaluates the quality of predicted trajectory at each time step. ADS, as defined in DriveArena\cite{yang2024drivearena}, combines PDMS with RC, where RC follows the same definition as the CARLA Leaderboard. 

The closed-loop performance of {\newmethodname} and end-to-end autonomous driving algorithms is summarized in Table \ref{tab:drivearena}. Notably, {\newmethodname} generates meta-actions, unlike other algorithms that directly output trajectories. We evaluated {\newmethodname} under three distinct configurations in DriveArena~\cite{yang2024drivearena}: 
\emph{\romannumeral 1)} {\newmethodname} + InternVL2 (Mem.), where InternVL2 serves as the VLM for scene understanding, and the system utilizes a memory bank;
\emph{\romannumeral 2)} {\newmethodname} + QwenVL, where QwenVL is used as the VLM for scene understanding, and the system operates without a memory bank;
\emph{\romannumeral 3)} {\newmethodname} + QwenVL (Mem.), where QwenVL is employed as the VLM, and the system incorporates a memory bank.

As demonstrated in Table \ref{tab:drivearena}, our knowledge-driven approach exhibits robust performance in high-fidelity simulation environments. In particular, the memory bank accumulated from the CARLA simulator works effectively when transferred to another simulator. This shows the strong generalization capability of our approach, enabling adaptation to diverse environmental conditions without excessive reliance on training data.

\subsection{Ablation Study}
\label{exp: Ablation}
We conduct extensive ablation studies about the number of few-shots, size of the memory bank, reflection mechanism, and accumulated experience in a closed-loop driving setup to demonstrate the generalization and continuous learning capabilities of our {\newmethodname}.

\subsubsection{Ablation of VLM}
\label{sec:vlm}

\begin{table}[t]
\vspace{3pt}
\captionsetup{font={small}}
  \centering
      \setlength{\tabcolsep}{1.1pt}
    \renewcommand{\arraystretch}{1.1}
    \caption{We assess the grounding and conversational capabilities of prominent VLMs using the validation set in Rank2Tell dataset~\cite{sachdeva2024rank2tell} and the CarlaSim dataset introduced in Section~\ref{sec:exp-data-vlm}
}
  \resizebox{\linewidth}{!}{
  \begin{tabular}{c|c|ccc|cc}
    \toprule
         Dataset  & Method & Precision$_\uparrow$  & Recall$_\uparrow$ & F1 Score$_\uparrow$ & ROUGE$_\uparrow$  & GPT Score$_\uparrow$\\
		\midrule
    \multirow{2}{*}{Rank2Tell} & llava-1.5-7b~\cite{zhang2023llava} & 28.49 &25.22 & 26.75 & 69.26 & 65.20 \\
    ~ & qwen-vl2-7b~\cite{bai2023qwen} & 46.70 & 37.37 & 41.52 & 70.23 & 66.59 \\
    \cite{sachdeva2024rank2tell} & internvl2-8b~\cite{chen2024internvl} & 45.67 & 36.94 & 40.84 & 62.71 & 63.60 \\
    \midrule
     \multirow{3}{*}{CarlaSim} & llava-1.5-7b~\cite{zhang2023llava} & 34.95 &32.00 & 33.41 & 78.04 & 58.09 \\
    ~ & qwen-vl-7b~\cite{bai2023qwen} & 51.41 & 47.14 & 49.18 & 83.24 & 63.01 \\
    ~ & internvl2-8b~\cite{chen2024internvl} & 53.55 & 44.80 & 48.79 & 73.77 & 58.13 \\
        \bottomrule

  \end{tabular}
  }
  \label{tab:vlm}
\end{table}

To select the most suitable VLM model for this task, we assessed the performance of various VLMs using the Rank2Tell (real environment) and CARLA (simulation environment) test sets. Specifically, we compared the performance of three models: LLaVA-1.5-7B~\cite{zhang2023llava}, Qwen-VL-7B~\cite{bai2023qwen}, and InternVL2-8B~\cite{chen2024internvl}. We utilized the Grounded Score, encompassing accuracy, recall, and F1 value, to assess the semantic alignment capabilities of the models. Additionally, we employed the Chat Score, which comprises the language score ROUGE and the GPT (gpt-4-turbo) score, to evaluate their question-answer reasoning abilities. Table~\ref{tab:vlm} presents the experimental results, indicating that Qwen-VL-7B and InternVL2-8B outperform the other models.

We further assessed the performance of Qwen-VL-7B and InternVL2-8B in a closed-loop setting. We conducted closed-loop tests using these two models on the CARLA Town05 short benchmark, and the experimental results are presented in the penultimate two rows of Table~\ref{tab:carla-leaderboard}. The experiment demonstrated the superior performance of QwenVL, thus, it was selected for subsequent experiments in this paper.

\begin{table}[th]
\setlength{\tabcolsep}{5.2pt}
\centering
    \renewcommand{\arraystretch}{1.1}
    \small
\caption{Precision@1 is evaluated for various scene tokens on nuScenes~\cite{caesar2020nuscenes}. For each scene token in the validation set, the most similar token from the training set is selected as the top-1 match. A pair of scene tokens is deemed a positive match for steering if the offset of the steering angle is within 0.04. The pair indicates a positive match for braking if the braking signals are concurrent.}
  \resizebox{\linewidth}{!}{
\begin{tabular}{c|c|cc} 
    \Xhline{0.5pt}
    \multirow{2}{*}{Method}
    &  \multirow{2}{*}{Ego state} &\multicolumn{2}{c}{Precision@1} \\
    
    ~ & ~ &  steer(\%)$\uparrow$ & brake(\%)$\uparrow$   \\
    \hline
    LeapAD~\cite{mei2024continuously} & & 60.39 & 76.62 \\
    Scene Encoder(pooling) &  & 82.69 & 83.55 \\
    Scene Encoder(pooling) & $\checkmark$ & \textbf{83.20} & \textbf{87.52} \\
    Scene Encoder(attention)& $\checkmark$ &  82.22 & 85.81 \\
    \Xhline{0.5pt}
\end{tabular}
}
\label{table:precision@1}
\end{table}

\subsubsection{Different type of scene token}
\label{sec:ab_scene_token}
To promote standardization, we assessed the performance of varying scene token representations on the nuScenes dataset~\cite{caesar2020nuscenes}. 
We utilize the nuScenes training set for model training, while the validation set serves as queries for retrieving data from the training set during evaluation. Table~\ref{table:precision@1} presents the precision@1 for each method, with our scene token approach demonstrating significantly superior performance compared to LeapAD. For evaluation, we retrieved the scene from the memory bank with the highest cosine similarity and utilized its associated driving steering and braking labels. 
The results revealed that the ``pooling + state'' setting yielded the best model performance, and thus, we adopted this setting as the default for the closed-loop experiments within this paper.

\begin{figure}[t]
    \centering
    \includegraphics[width=\linewidth]{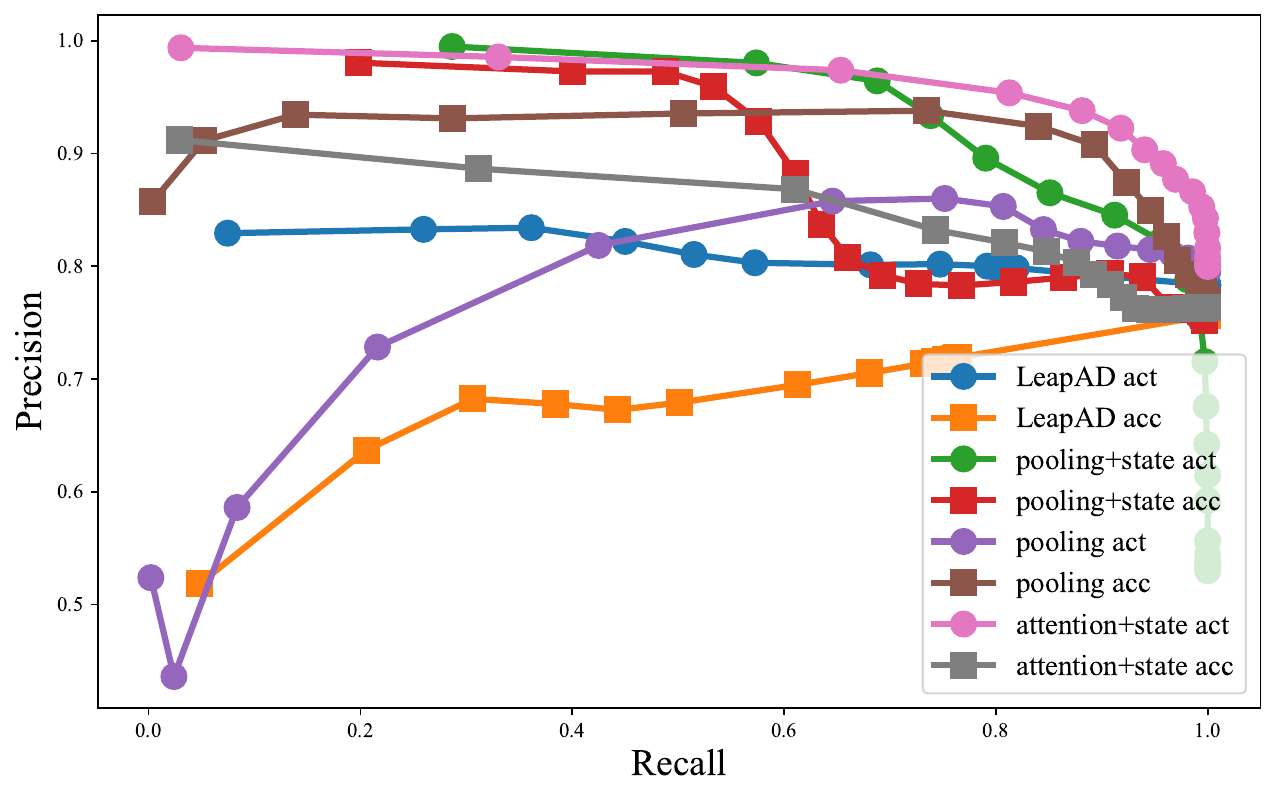}
    \caption{Precision-Recall curves on nuScenes~\cite{caesar2020nuscenes} dataset}
    \label{fig:pr-curve}
\end{figure}

To assess the performance of varying scene token representations more comprehensively, we present the precision-recall curves for different methods in Figure~\ref{fig:pr-curve}. The legend labels the method, respective settings, and classification space. It is evident from the figure that LeapAD~\cite{mei2024continuously} exhibits the poorest performance among the compared methods. In contrast, the ``pooling + state'' and ``attention+state'' methods demonstrate a more balanced performance, with ``pooling+state'' exhibiting superior results in the ACC space, which is particularly crucial for closed-loop experiments.

\begin{figure*}
    \centering
    \includegraphics[width=\linewidth]{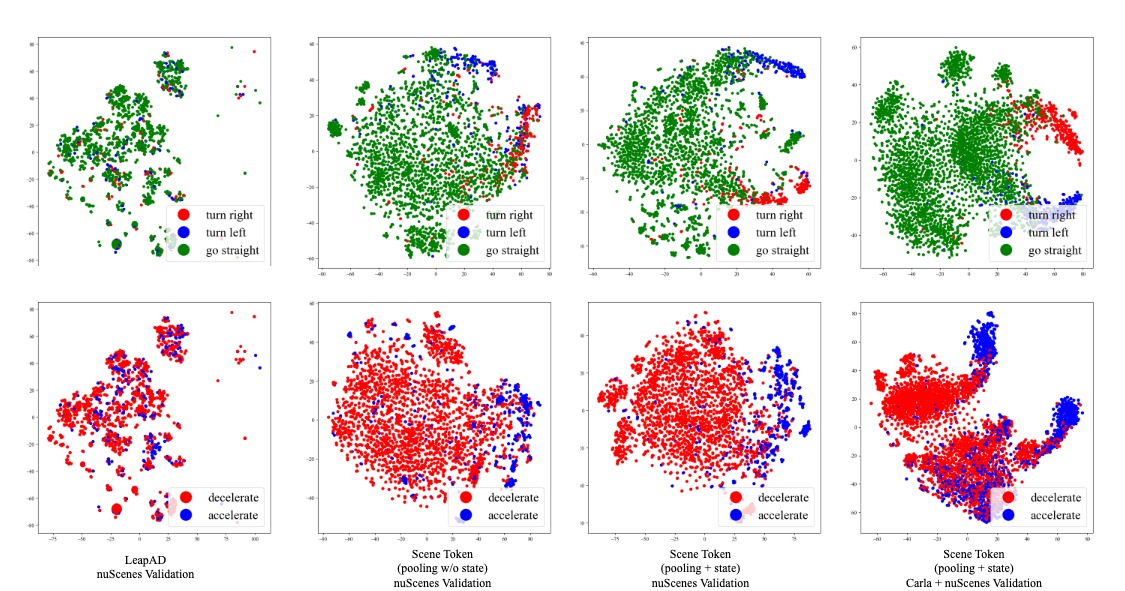}
    \caption{
    t-SNE~\cite{hinton2008visualizing} visualization of extracted features. The first row illustrates the relationship between $\boldsymbol g_{act}$ and the steering, whereas the second row depicts the correlation between $\boldsymbol g_{acc}$ and the brake. The category of the scene token is indicated at the bottom of each respective column.
    } %
    \label{fig:tsne}
\end{figure*}

To better illustrate the enhancements attributed to the scene token, we employ t-SNE~\cite{hinton2008visualizing} to visualize its impact on clustering different scene token representations. In the first column, Figure~\ref{fig:tsne} reveals that the scene token utilized by {\methodname} fails to form distinct clusters associated with possible driving actions. In contrast, the second column highlights our proposed scene token method's successful differentiation of steering and acceleration/deceleration patterns, even without ego state input. The inclusion of ego-state information further enhances the clustering effect. Notably, the fourth column demonstrates the effectiveness of our proposed scene token extraction technique on mixed nuScenes and CARLA data, yielding favorable clustering results.

Furthermore, Rows 13 and 14 of Table~\ref{tab:carla-leaderboard} show the results of closed-loop experiments conducted by {\newmethodname} using OpenAI Embedding and our scene token for sample retrieval, respectively. It is evident that utilizing our scene token yields superior outcomes.

\subsubsection{Ablation on the number of few-shot}
We conducted an ablation experiment on few-shot prompting in our previous version~\cite{mei2024continuously}. To determine if the same outcome would occur after updating the representation of the scene token, we replicated the experiment. The tests were conducted on the Town05 Short benchmark, comprising 9K samples randomly collected from the memory bank introduced in Section~\ref{sec:dataforfast}. As illustrated in Figure~\ref{fig:few-shot memsize}, the results demonstrate that our new scene token effectively retrieves relevant samples and guides decision-making. Due to our VLM updates, {\newmethodname} outperforms other camera-only methods, even in a zero-shot setting. A notable performance enhancement is observed when advancing from zero-shot to three-shot scenarios, underscoring the advantages of leveraging experience stored in memory and the efficacy of the few-shot strategy.

\subsubsection{The impact of memory sizes}
\label{sec: size of memory}
\begin{figure}[t]
    \centering
    \includegraphics[width=\linewidth]{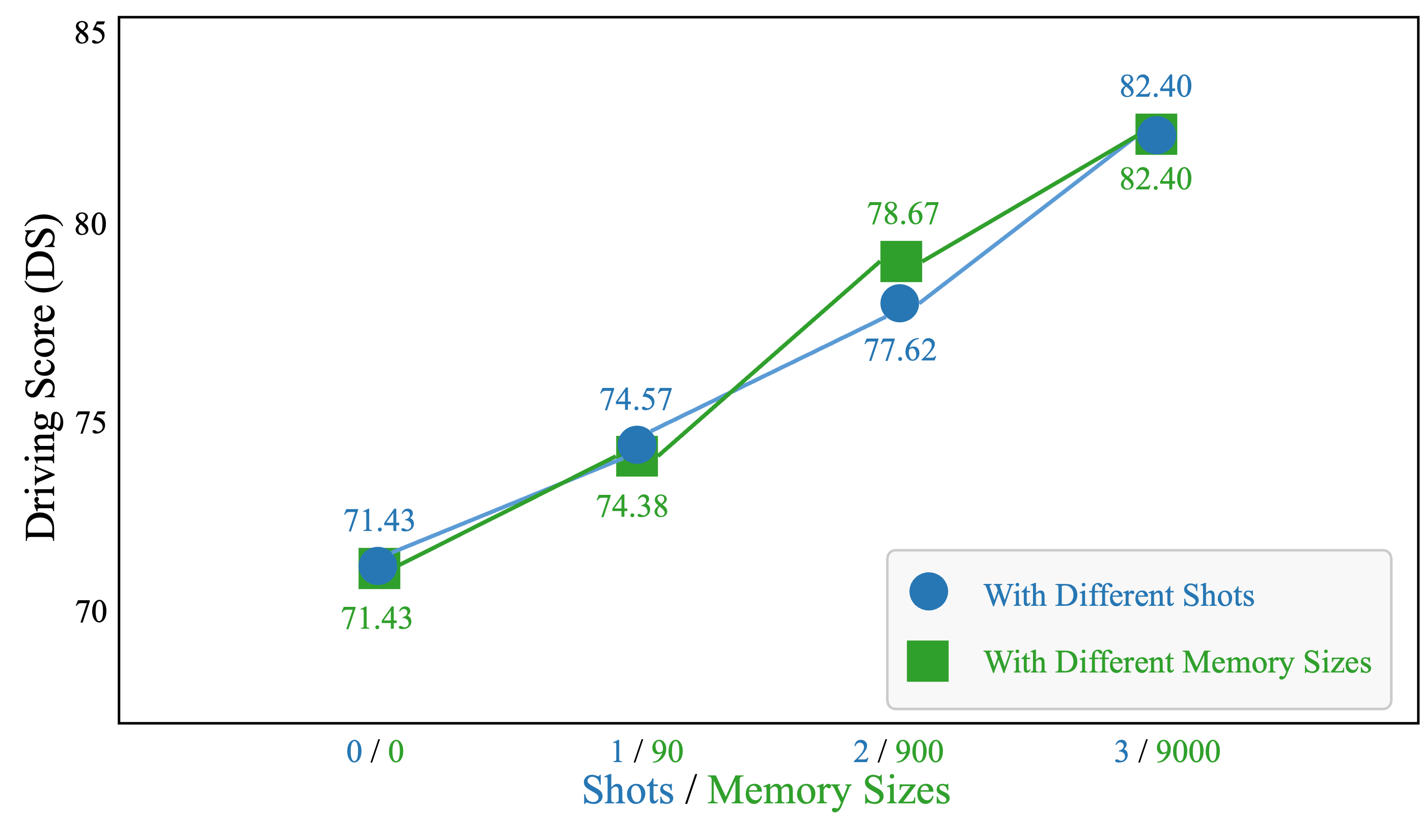}
    \caption{The illustration for ablation studies of few-shot and memory size. 
    }
    \label{fig:few-shot memsize}
    \vspace{-15pt}
\end{figure}

Memory capacity plays a pivotal role in knowledge-driven systems by aggregating critical experiences. Through comprehensive ablation studies, we demonstrate that larger memory capacities lead to enhanced system performance. We conduct closed-loop evaluations using three different memory sizes (9000, 900, and 90 experiences), with samples randomly drawn from the memory bank described in Section~\ref{sec:dataforfast}. As illustrated in Figure~\ref{fig:few-shot memsize}, the quantitative results reveal a consistent performance improvement with increasing memory size. This underscores the continuous learning ability of {\newmethodname} and establishes a direct correlation between model performance and experience accumulation.

\subsubsection{Effectiveness of the reflection mechanism}
In our prior work~\cite{mei2024continuously}, we sampled nine sequences with scores below 50 on the Town05 short benchmark for reflection experiments. Specifically, we used a 3-shot and 900-experience memory bank setting and found that the scores of most sequences improved with additional reflection rounds, effectively demonstrating the validity of our designed reflection mechanism. We conducted additional ablation experiments to further substantiate this mechanism's robustness. As evidenced by experiments 14 and 15 in Table~\ref{tab:carla-leaderboard}, with all other parameters held constant, we observed a DS improvement of 4.41 and an RC enhancement of 0.11 after four reflection rounds.

\subsubsection{Generalization of accumulated knowledge}
\label{ab:generalization}
Our previous research demonstrated the robust transferability of memory banks across different towns in CARLA. In this section, we further investigate experience transfer from simulation environments to high-fidelity driving scenarios. As shown in the penultimate row of Table~\ref{tab:drivearena}, we first conducted closed-loop evaluations of {\newmethodname} in DriveArena without utilizing any memory bank. Subsequently, we incorporated 18.1k memories accumulated from CARLA, with results presented in the last row of Table~\ref{tab:drivearena}. The substantial improvement of 3.59 points in average PDMS demonstrates that experiences acquired in simulation environments can effectively transfer not only between different maps but also to real-world driving scenarios. This cross-domain performance enhancement, which surpasses the capabilities of existing methods, validates the superiority and robust transferability of our approach.

\section{Conclusion}
In this paper, we propose a novel, human-like, knowledge-driven autonomous driving framework that employs a VLM to focus on critical objects in the driving environment. Our framework mimics the way human eyes perceive and prioritize information. We have designed a dual-process decision-making module to emulate the learning process of human drivers. Additionally, we have incorporated reflective mechanisms and cumulative memory banks to enable continuous self-improvement of algorithms. 
We introduce an efficient scene token to represent the scene relevant to driving actions. This scene token retrieves similar samples to guide the current decision. Through our evaluations, our method demonstrates impressive closed-loop performance with reduced training data requirements. 
Specifically, we achieved performance that is near state-of-the-art (SOTA) using only 1/73 of the data. Additionally, we improved the driving scores on the Town05 short and long benchmarks by 5.3\% and 42.6\%, respectively, compared to LeapAD~\cite{mei2024continuously}, and attained the best performance in DriveArena~\cite{yang2024drivearena}.
Furthermore, extensive ablation studies validate the efficiency of the proposed modules and highlight the continuous learning ability and transferability of knowledge within our system.
\section*{Appendix}

\subsection{Low-level Control}
\label{sec:low-level-control}
Our dual-process decision-making module outputs meta-actions (e.g., ``AC'', ``DC'', ``LCL'', ``LCR'', ``IDLE'', ``STOP''), which serve as inputs for trajectory generation. A PID controller then tracks these trajectories to compute the final control signals, such as steering, throttle, and brake.

\subsubsection{Planned Waypoints}
To enhance the effectiveness of route tracking, we addressed the issue of the potentially large gap (up to several dozen meters) between the default waypoints provided by CARLA. By leveraging high-definition maps, we converted these sparse waypoints into dense path points, spaced 1 meter apart, thus creating a precise reference path for the vehicle. Furthermore, when needed, we incorporated information about adjacent lanes along this reference path to support lane-changing operations, such as overtaking or obstacle avoidance. This enhancement not only allows the vehicle to deviate from the navigation waypoints as required temporarily but also significantly improves the flexibility and execution capability of the control system.

Combining detailed lane data, our controller uses a planner provided by LimSim~\cite{wen2023limsim} to generate trajectories for the next 5 seconds, ensuring that the vehicle travels on the desired path. The actions ``AC'' and ``DC'' determine the target state by calculating the target acceleration from the current acceleration, establishing the speed and position 5 seconds ahead. The actions ``LCL'' and ``LCR'' use a spatio-temporal sampling strategy to sample target positions within the target lane and the speed range to determine the target state. These target states are inputted into the trajectory optimizer in Frenét coordinate~\cite{wen2023bringing}, generating quintic polynomial trajectories. As a result, the trajectory is optimized and selected based on cost factors such as smoothness, speed matching, acceleration, jerk, and obstacle avoidance, resulting in the optimal trajectory. 

Our approach can utilize alternative methods rather than relying on high-definition maps. Techniques like those in DriveCot~\cite{wang2024drivecot} and TransFuser~\cite{chitta2022transfuser} leverage distinct neural networks to predict future paths from camera images and sparse navigation data, seamlessly aligning with controller design while preserving our core methodology.

\subsubsection{PID Controller}
The controller selects trajectory points, including target speeds and target waypoints, from the sequence planned by the LimSim planner based on the vehicle's current speed: at lower speeds, it chooses points further along in the sequence, and at higher speeds, it selects earlier points to ensure effective tracking control.

For controlling the vehicle's motion, two independent PID controllers are employed, as outlined in previous works~\cite{chitta2021neat, chitta2022transfuser, wu2022trajectory, jia2023think, wang2024drivecot}. One controller manages the longitudinal motion for throttle and braking to track the target speed, while the other handles the lateral motion for steering to follow the designated waypoint. The longitudinal PID controller, tuned with gains $K_P =1.95$, $K_I = 0.05$, and $K_D = 0.2$, adjusts the throttle and brake based on the current and target speeds, using a 10-frame buffer. The lateral PID controller, $K_P = 1.0$, $K_I = 0.05$, and $K_D = 0.0$, calculates the steering value from the heading angle difference to the next waypoint using a 10-frame buffer.

\subsection{Prompt Details}
\label{sec:prompt}
\begin{figure}[htbp]
    \centering
    \includegraphics[width=\linewidth]{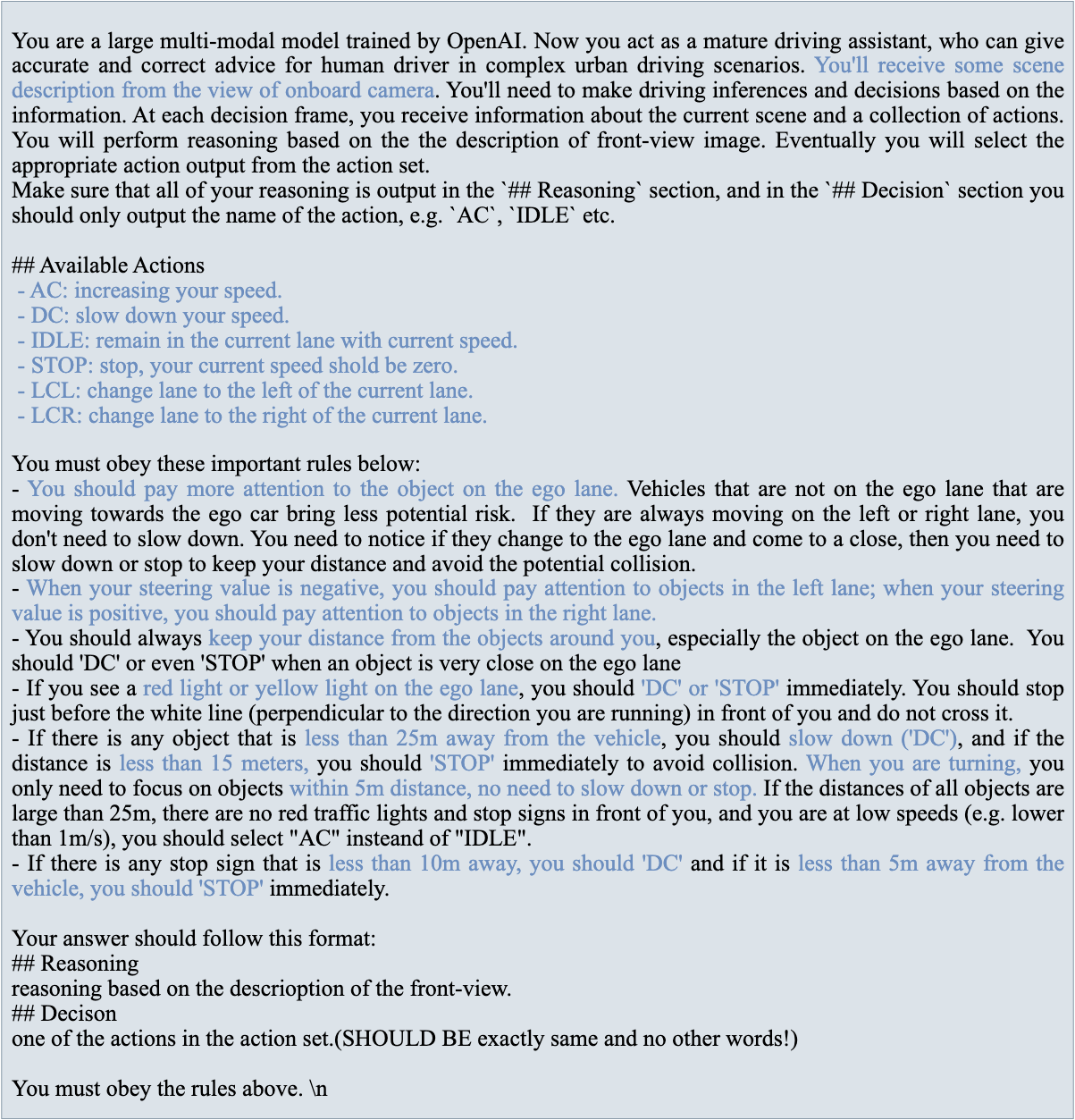}
    \caption{System prompt for \slowsystem}
    \label{fig:system prompt}
\end{figure}

\begin{figure}[htbp]
    \centering
    \includegraphics[width=\linewidth]{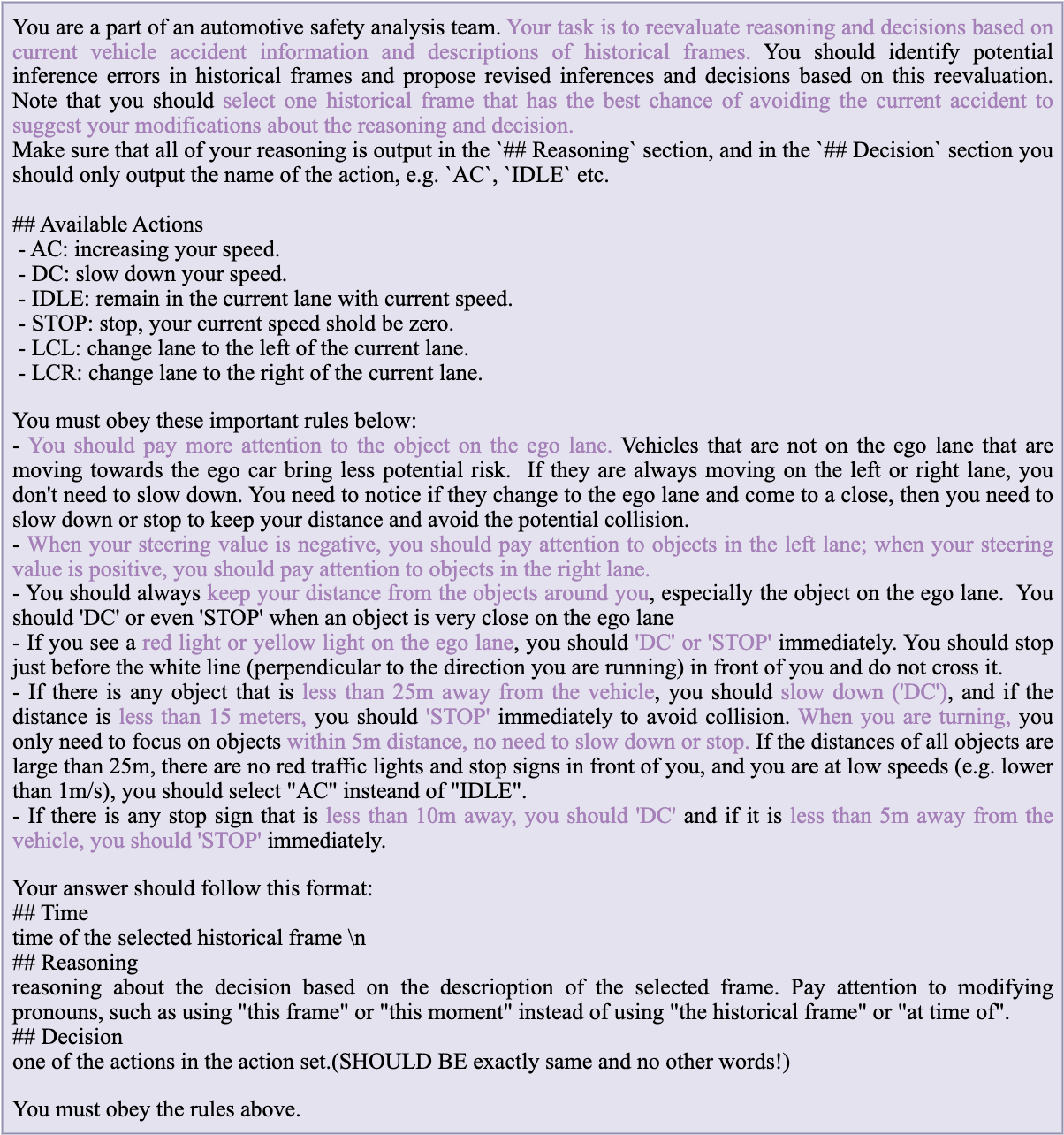}  
    \caption{System prompt in the reflection mechanism}  
    \label{fig:reflection prompt}
\end{figure}

\begin{figure}[htbp]
    \centering
    \includegraphics[width=\linewidth]{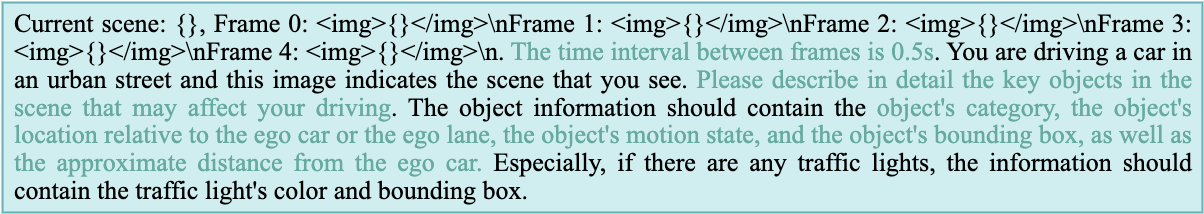}  
        \caption{VLM system prompt} 
        \label{fig:vlm prompt}
\end{figure}

We detail the system prompt (Figure~\ref{fig:system prompt}) used by our {\slowsystem} for experience accumulation in a closed-loop environment. This prompt includes task definitions, meta-actions, traffic rule adherence, and output format. Additionally, Figure~\ref{fig:reflection prompt} shows the reflection prompt, which incorporates key elements from the previous prompt and criteria for detecting potential errors in historical frames. We also describe the VLM prompts (Figure~\ref{fig:vlm prompt}) used to identify critical objects in traffic scenes.

\subsection{Visualization Cases}
\label{sec:visualization}
\subsubsection{case of scene token}
\begin{figure}[htbp]
    \centering
    \includegraphics[width=\linewidth]{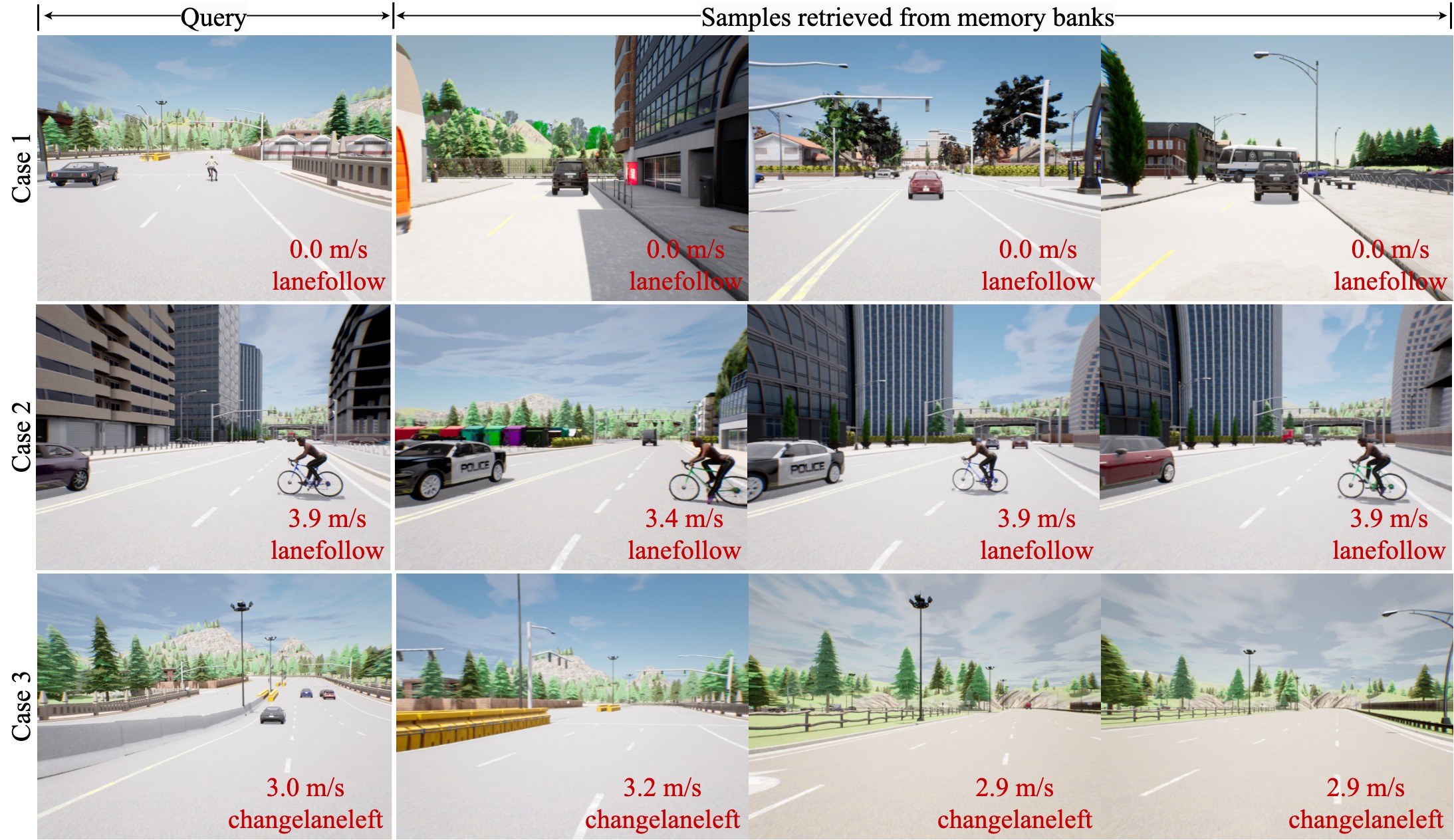}
    \caption{Retrieval samples in the CARLA closed-loop experiment}
    \label{fig:token_case_carla}
\end{figure}
\begin{figure}[htbp]
    \centering
    \includegraphics[width=\linewidth]{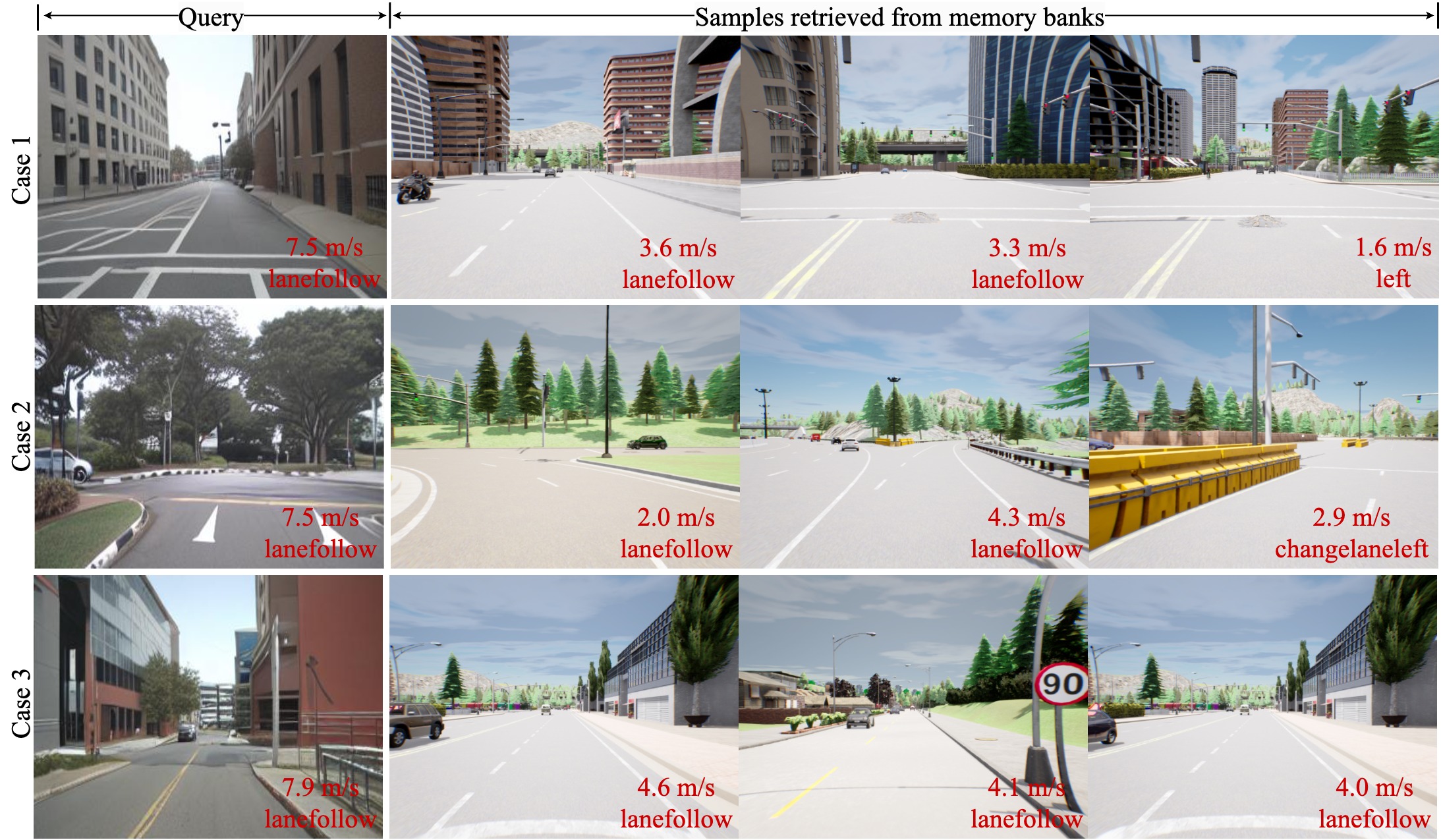}
    \caption{Retrieval samples in the DriveArena closed-loop experiment}
    \label{fig:token_case_arena}
\end{figure}
In our closed-loop experiments, we selected several cases about the samples retrieved from the memory bank using our scene token. Figure~\ref{fig:token_case_carla} illustrates examples from the Town05 short benchmark—notably, the samples retrieved from the memory bank exhibit similar scenario characteristics and traffic conditions. The ego state of the query closely aligns with that of the samples. Figure~\ref{fig:token_case_arena} presents cases from DriveArena.
There is a discernible correlation between the query and the samples. For instance, in Case 1, the query pertains to a scene with crippled traffic lights, and all samples include traffic elements. Case 2 focuses on intersections, while Case 3 involves oncoming traffic in the opposite direction. However, due to the significant data gap between the query and the memory bank, the retrieval quality is notably inferior to that depicted in Figure~\ref{fig:token_case_carla}.

\subsubsection{Close-loop case in DriveArena}
\begin{figure}[htbp]
    \centering
    \includegraphics[width=\linewidth]{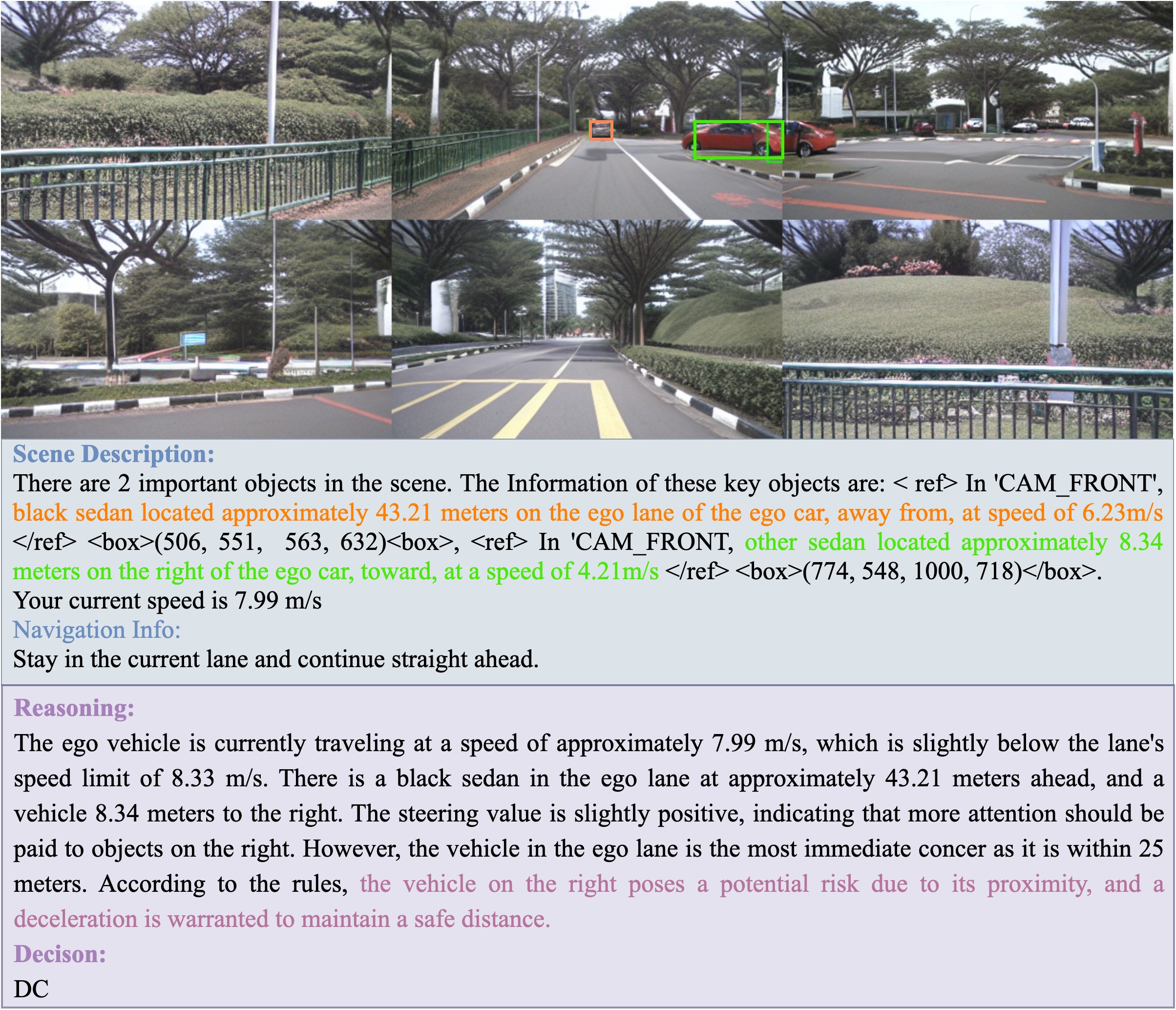}
    \caption{A case where a vehicle suddenly cut in.}
    \label{fig:arena_case}
\end{figure}

Figure \ref{fig:arena_case} illustrates a closed-loop case in DriveArena. It depicts a scenario where the ego car is driving, aligned with its designated lane. Suddenly, a red car cuts in from the right, and the VLM indicates that another sedan is approaching from the right as well. In response, the {\fastsystem} decides to decelerate to maintain a safe distance from the approaching vehicle.
\begin{figure}[htbp]
    \centering
    \includegraphics[width=\linewidth]{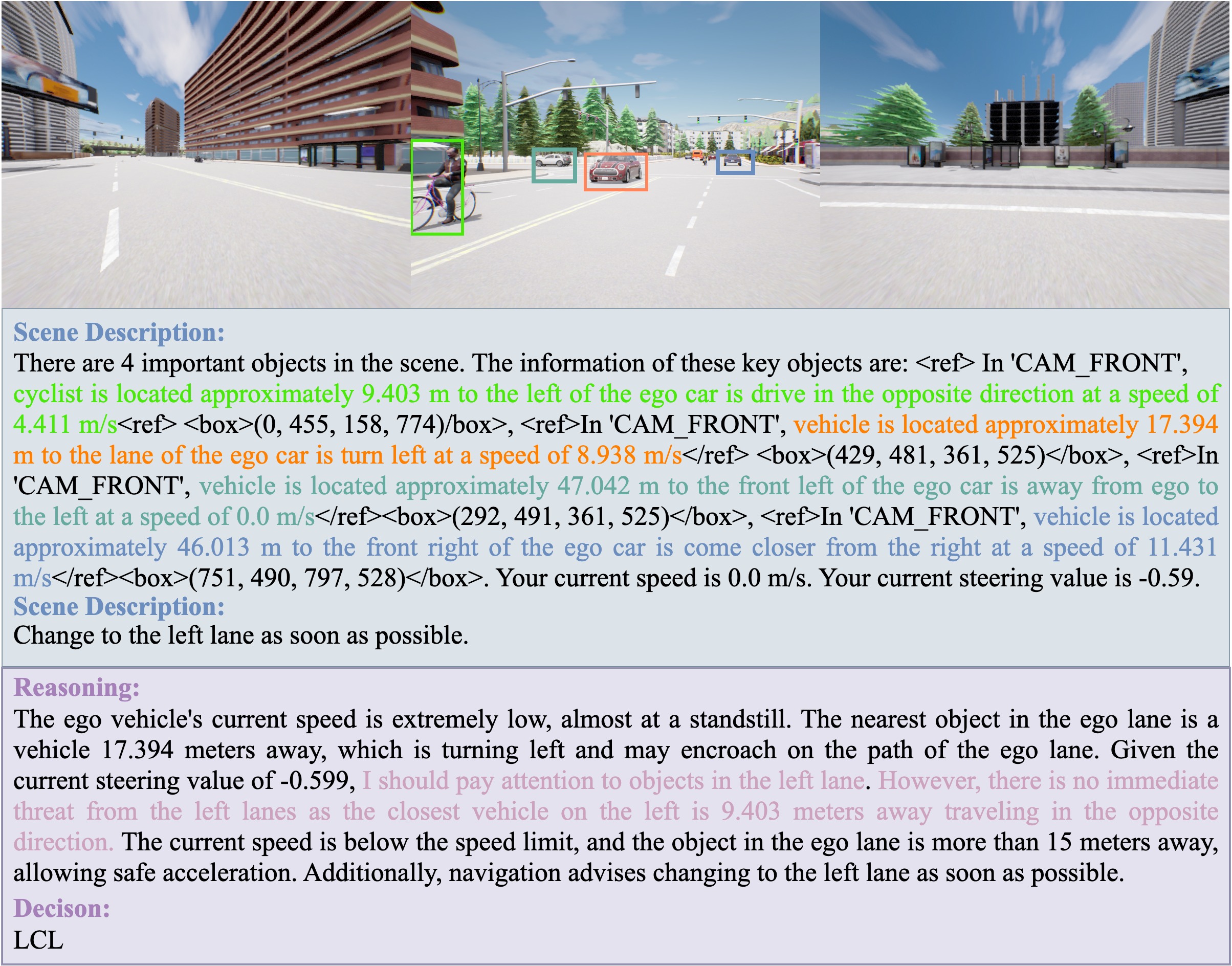}
    \caption{A case of ``LCL''.}
    \label{fig:carla_case}
\end{figure}

\subsubsection{Close-loop case in CARLA}
Figure \ref{fig:carla_case} illustrates a closed-loop case in CARLA, where the ego car navigates through a city with heavy traffic. The navigation system advises the car to switch to the left lane. As depicted in VLM, the majority of the vehicles are at a safe distance from the ego car. The closest cyclist is traveling in the opposite direction on the left, but {\fastsystem} assesses that changing lanes presents no risk and makes the decision to ``LCL''.

\subsubsection{Cases of reflection mechanism}
\label{sec:reflection}

\begin{figure}[htbp]
    \centering
    \includegraphics[width=\linewidth]{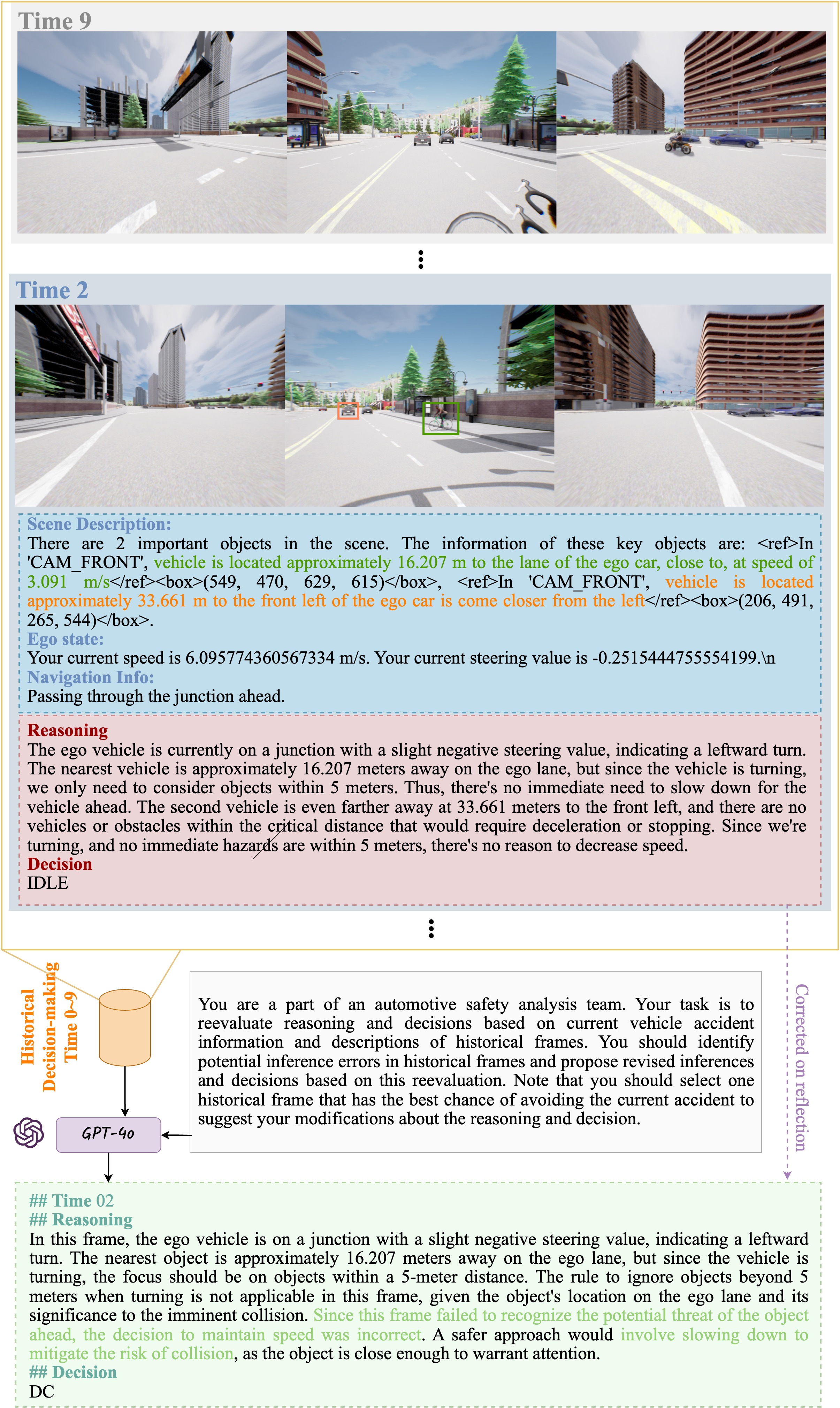}
    \caption{Case study for reflection mechanism.}
    \label{fig:reflectioncase}
\end{figure}

As mentioned in Section \ref{method:slow}, we utilize the {\slowsystem} to reflect on traffic accidents, enhancing the system's capabilities through continuous improvement. Expressly, we set the maximum length of the memory queue $\boldsymbol{Q}$ to $m=10$ at a frequency of 1 Hz. As illustrated in Figure~\ref{fig:reflectioncase}, when the {\fastsystem} collided with a cyclist at time step 9, we input the accident type $O$, memory queue $\boldsymbol{Q}$, and the reflection system prompt (Figure \ref{fig:reflection prompt}) into the {\slowsystem} to rectify past errors. The output from the {\slowsystem} suggests that the car should ``DC'' at time step 2 instead of ``IDLE'' to avoid the risk of collision.

{\small
\bibliographystyle{ieeetr}  
\bibliography{ref}
}

\end{document}